%% file: main.tex
\documentclass[10pt,twocolumn,letterpaper]{article}

\usepackage{cvpr}              
\usepackage[accsupp]{axessibility}

\usepackage{graphicx}
\usepackage{amsmath}
\usepackage{amssymb}
\usepackage{booktabs}

\usepackage{multirow}
\usepackage{algorithm}
\usepackage[noend]{algpseudocode}
\usepackage{sidecap}
\usepackage[dvipsnames]{xcolor}
\usepackage{pifont}
\newcommand{\cmark}{{\color{ForestGreen}\ding{51}}}
\newcommand{\xmark}{{\color{BrickRed}\ding{55}}}

\newcommand{\ourmodel}{NeuralField-LDM}
\newcommand{\ourmodelsrt}{NF-LDM}

\usepackage[pagebackref,breaklinks,colorlinks]{hyperref}

\usepackage[capitalize]{cleveref}
\crefname{section}{Sec.}{Secs.}
\Crefname{section}{Section}{Sections}
\Crefname{table}{Table}{Tables}
\crefname{table}{Tab.}{Tabs.}

\def\cvprPaperID{7140} 
\def\confName{CVPR}
\def\confYear{2023}

\begin{document}

\title{NeuralField-LDM: Scene Generation with Hierarchical Latent Diffusion Models}
\author{
    Seung Wook Kim\textsuperscript{1,2,3*}
    \quad Bradley Brown\textsuperscript{1,5*\dag}
    \quad\enspace Kangxue Yin\textsuperscript{1}
    \quad Karsten Kreis\textsuperscript{1}
    \quad\enspace Katja Schwarz\textsuperscript{6\dag}
    \\
    \quad\quad Daiqing Li\textsuperscript{1}
    \quad\quad Robin Rombach\textsuperscript{7\dag}
    \quad\quad Antonio Torralba\textsuperscript{4}    
    \quad\quad Sanja Fidler\textsuperscript{1,2,3} \\
    \small{
        \textsuperscript{1}NVIDIA 
        \quad \textsuperscript{2}University of Toronto 
        \enspace \textsuperscript{3}Vector Institute 
        \quad \textsuperscript{4} CSAIL, MIT 
        \quad \textsuperscript{5}University of Waterloo
    }\\
    \small{
        \textsuperscript{6}University of Tübingen, Tübingen AI Center
        \quad \textsuperscript{7}LMU Munich
    }\\
    \url{https://research.nvidia.com/labs/toronto-ai/NFLDM}\\
}

\twocolumn[{%
\renewcommand\twocolumn[1][]{#1}%
\vspace{-6mm}
\maketitle
\begin{center}
\vspace{-8.5mm}
    \centering
    \captionsetup{type=figure}
    \includegraphics[width=1\textwidth]{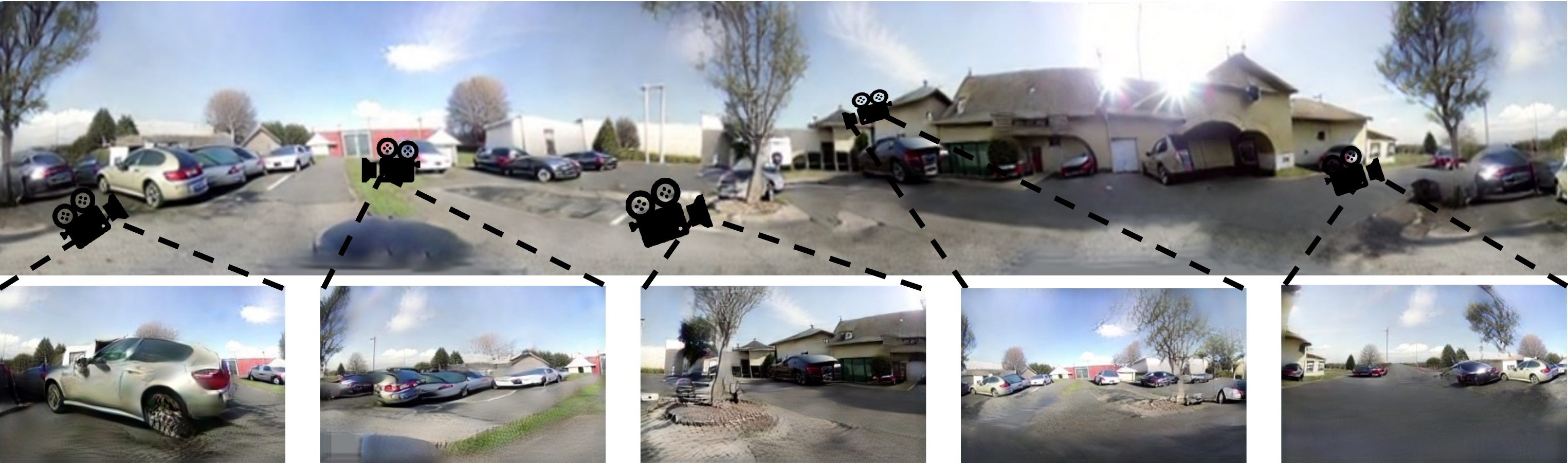}
    \vspace{-7mm}
    \captionof{figure}{We introduce {\ourmodel}, a generative model for complex open-world 3D scenes. This figure contains a panorama constructed from {\ourmodel}'s generated scene. We visualize different parts of the scene by placing cameras on them.}
\end{center}%
}]


\begin{abstract}

Automatically generating high-quality real world 3D scenes is of enormous interest for applications such as virtual reality and robotics simulation.
Towards this goal, we introduce \ourmodel, a generative model capable of synthesizing complex 3D environments.
We leverage Latent Diffusion Models that have been successfully utilized for efficient high-quality 2D content creation.
We first train a scene auto-encoder to express a set of image and pose pairs as a neural field, represented as density and feature voxel grids that can be projected to produce novel views of the scene.
To further compress this representation, we train a latent-autoencoder that maps the voxel grids to a set of latent representations.
A hierarchical diffusion model is then fit to the latents to complete the scene generation pipeline. 
We achieve a substantial improvement over existing state-of-the-art scene generation models.
Additionally, we show how {\ourmodel} can be used for a variety of 3D content creation applications, including conditional scene generation, scene inpainting and scene style manipulation.  

{\let\thefootnote\relax\footnote{{\textsuperscript{*}Equal contribution.}}}
{\let\thefootnote\relax\footnote{{\textsuperscript{\dag}Work done during an internship at NVIDIA.}}}
\end{abstract}


\input{sections/introduction}
\input{sections/related_works}
\input{sections/method}

\input{sections/experiments}

\input{sections/conclusion}

\clearpage
{\small
\bibliographystyle{ieee_fullname}
\bibliography{bibliography}
}

\clearpage
\input{sections/supple_embed}
\end{document}

%% file: sections/introduction.tex
\vspace{-4mm}
\section{Introduction}
\label{sec:intro}
There has been increasing interest in modelling 3D real-world scenes for use in virtual reality, game design, digital twin creation and more.
However, designing 3D worlds by hand is a challenging and time-consuming process, requiring 3D modeling expertise and artistic talent.
Recently, we have seen success in automating 3D content creation via 3D generative models that output individual object assets \cite{gao2022get3d,poole2022dreamfusion,Zeng2022ARXIV}.
Although a great step forward, automating the generation of real-world scenes remains an important open problem and would unlock many applications ranging from scalably generating a diverse array of environments for training AI agents (\eg autonomous vehicles) to the design of realistic open-world video games.
In this work, we take a step towards this goal with {\ourmodel} (\ourmodelsrt), a generative model capable of synthesizing complex real-world 3D scenes.
{\ourmodelsrt} is trained on a collection of posed camera images and depth measurements which are easier to obtain than explicit ground-truth 3D data, offering a scalable way to synthesize 3D scenes.

Recent approaches \cite{DeVries2021ICCV,Chan2022CVPR, Bautista2022ARXIV} tackle the same problem of generating 3D scenes, albeit on less complex data. 
In \cite{DeVries2021ICCV,Chan2022CVPR}, a latent distribution is mapped to a set of scenes using adversarial training, and in GAUDI~\cite{Bautista2022ARXIV}, a denoising diffusion model is fit to a set of scene latents learned using an auto-decoder.
These models all have an inherent weakness of attempting to capture the entire scene into a single vector that conditions a neural radiance field. 
In practice, we find that this limits the ability to fit complex scene distributions.

Recently, diffusion models have emerged as a very powerful class of generative models, capable of generating high-quality images, point clouds and videos \cite{Rombach2022CVPR,Ramesh2022ARXIV, Zhou2021ICCV,Luo2021CVPR,Zeng2022ARXIV,ho2022video, harveryflexvideo}.
Yet, due to the nature of our task, where image data must be mapped to a shared 3D scene without an explicit ground truth 3D representation, straightforward approaches fitting a diffusion model directly to data are infeasible.


In {\ourmodel}, we learn to model scenes using a three-stage pipeline.
First, we learn an auto-encoder that encodes scenes into a neural field, represented as density and feature voxel grids. 
Inspired by the success of latent diffusion models for images \cite{Rombach2022CVPR}, we learn to model the distribution of our scene voxels in latent space to focus the generative capacity on core parts of the scene and not the extraneous details captured by our voxel auto-encoders.
Specifically, a latent-autoencoder decomposes the scene voxels into a 3D coarse, 2D fine and 1D global latent.
Hierarchichal diffusion models are then trained on the tri-latent representation to generate novel 3D scenes.
We show how {\ourmodelsrt} enables applications such as scene editing, birds-eye view conditional generation and style adaptation. 
Finally, we demonstrate how score distillation~\cite{poole2022dreamfusion} can be used to optimize the quality of generated neural fields, allowing us to leverage the representations learned from state-of-the-art image diffusion models that have been exposed to orders of magnitude more data.

Our contributions are: 
	1) We introduce {\ourmodelsrt}, a hierarchical diffusion model capable of generating complex open-world 3D scenes and achieving state of the art scene generation results on four challenging datasets.
	2) We extend {\ourmodelsrt} to semantic birds-eye view conditional scene generation, style modification and 3D scene editing.

%% file: sections/related_works.tex
\vspace{-2mm}
\section{Related Work}
\label{sec:related_work}
\paragraph{2D Generative Models}
In past years, generative adversarial networks (GANs)~\cite{Goodfellow2014NIPS,Mescheder2018ICML,Brock2019ICLR,Karras2020CVPRa,Sauer2022ARXIV} and likelihood-based approaches~\cite{Kingma2014ICLR,Rezende2014ICML,Vahdat2020NIPS,Razavi2019NIPS} enabled high-resolution photorealistic image synthesis. Due to their 
quality, GANs are used in a multitude of downstream applications ranging from steerable content creation~\cite{Liao2020CVPR,Ling2021NIPS,li2022ARXIV,
Shen2020CVPR,ZhangICLR2021,Kim2022_PMGAN} to data driven simulation~\cite{kim2020learning,Kim2021CVPR,Kar2019ICCV,li2022ARXIV}. 
Recently, autoregressive models and score-based models, e.g. diffusion models, demonstrate better distribution coverage while preserving high sample quality~\cite{Ho2020NIPS,Ho2022MLRes,Nichol2022ICML,
Dockhorn2022ICLR,Ramesh2022ARXIV,Dhariwal2021NIPS,Vahdat2021NIPS,Rombach2021ICCV,Rombach2022CVPR,Esser2021CVPR}. Since evaluation and optimization of these approaches in pixel space is computationally expensive,~\cite{Vahdat2021NIPS,Rombach2022CVPR} apply them to latent space, achieving state-of-the-art image synthesis at megapixel resolution. 
As our approach operates on 3D scenes, 
computational efficiency is crucial. Hence, we build upon~\cite{Rombach2022CVPR} and train our model in latent space.

\vspace{-4mm}
\paragraph{Novel View Synthesis}
In their seminal work~\cite{Mildenhall2020ECCV}, Mildenhall et al. introduce Neural Radiance Fields (NeRF) as a powerful 3D representation. 
PixelNeRF~\cite{Yu2021CVPR} and IBRNet~\cite{Wang2021CVPR} propose to condition NeRF on aggregated features from multiple views to enable novel view synthesis from a sparse set of views. 
Another line of works scale NeRF to large-scale indoor and outdoor scenes~\cite{Zhang2020ARXIVc,Zhang2022CVPR,Martin-Brualla2021CVPR,Rematas2022CVPR}. 
Recently, Nerfusion~\cite{Zhang2022CVPR} predicts local radiance fields and fuses them into a scene representation using a recurrent neural network. Similarly, we construct a latent scene representation by aggregating features across multiple views. 
Different from the aforementioned methods, our approach is a generative model capable of synthesizing novel scenes.


\vspace{-4mm}
\paragraph{3D Diffusion Models}
A few recent works propose to apply denoising diffusion models (DDM)~\cite{song2020score,Ho2020NIPS,Ho2022MLRes} on point clouds for 3D shape generation~\cite{Zhou2021ICCV,Luo2021CVPR,Zeng2022ARXIV}. While PVD~\cite{Zhou2021ICCV} trains on point clouds directly, DPM~\cite{Luo2021CVPR} and LION~\cite{Zeng2022ARXIV} use a shape latent variable. 
Similar to LION, we design a hierarchical model by training separate conditional DDMs. 
However, our approach generates both texture and geometry of a scene without needing 3D ground truth as supervision.

\vspace{-4mm}
\paragraph{3D-Aware Generative Models}
3D-aware generative models synthesize images while providing explicit control over the camera pose and potentially other scene properties, like object shape and appearance.
SGAM~\cite{shen2022sgam} generates a 3D scene by autoregressively generating sensor data and building a 3D map. 
Several previous approaches generate NeRFs of single objects with conditional coordinate-based MLPs
~\cite{Schwarz2020NEURIPS,Chan2020CVPR,Niemeyer2021CVPR}. GSN~\cite{DeVries2021ICCV} conditions a coordinate-based MLP on a ``floor plan", i.e. a 2D feature map, to model more complex indoor scenes. EG3D~\cite{Chan2022CVPR} and VoxGRAF~\cite{Schwarz2022NIPS} use convolutional backbones to generate 3D representations. All of these approaches rely on adversarial training. Instead, we train a DDM on voxels in latent space. The work closest to ours is GAUDI~\cite{Bautista2022ARXIV}, which first trains an auto-decoder and subsequently trains a DDM on the learned latent codes. Instead of using a global latent code, we encode scenes onto voxel grids and train a hierarchical DDM to optimally combine global and local features.

%% file: sections/method.tex
\section{\ourmodel}
\label{sec:method}

\begin{figure*}[!thb]
\vspace{-1mm}
  \centering
\includegraphics[width=0.88\textwidth]{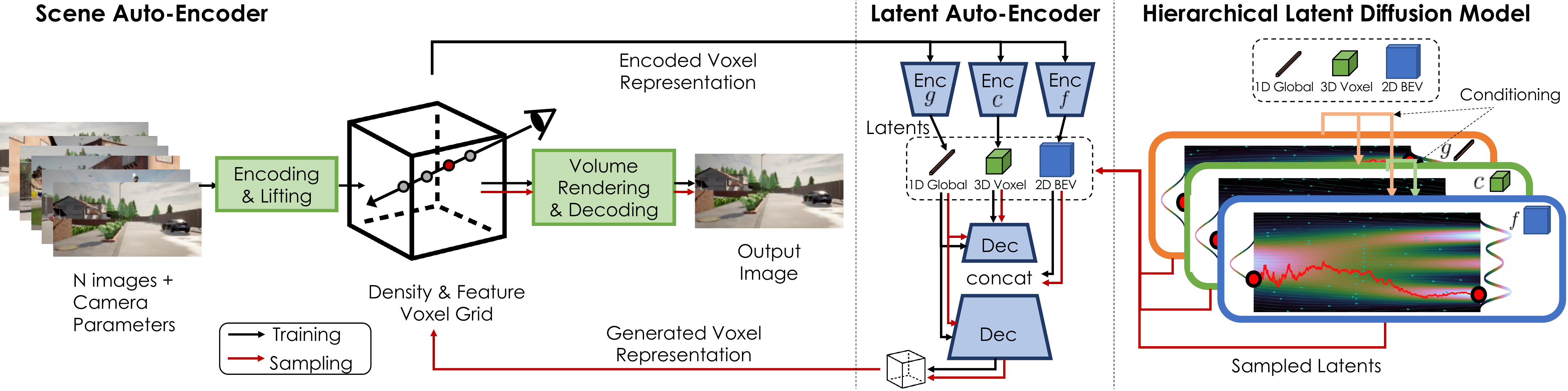}
\vspace{-3mm}
   \caption{ 
   \textbf{Overview of \ourmodel}. 
   We first encode RGB images with camera poses into a neural field represented by density and feature voxel grids. We compress the neural field into smaller latent spaces and fit a hierarchical latent diffusion model on the latent space. Sampled latents can then be decoded into a neural field that can be rendered into a given viewpoint. 
    }
\label{fig:pipeline}
\vspace{-2.5mm}
\end{figure*}

Our objective is to train a generative model to synthesize 3D scenes that can be rendered to any viewpoint.
We assume access to a dataset 
$\{(i, \kappa, \rho)\}_{1..N}$ which consists of $N$ RGB images $i$  and their camera poses $\kappa$, along with a depth measurement $\rho$ that can be either sparse (\eg Lidar points) or dense.
The generative model must learn to model both the texture and geometry distributions of the dataset in 3D by learning solely from the sensor observations, which is a highly non-trivial problem.

Past work typically tackles this problem with a generative adversarial network (GAN) framework~\cite{Schwarz2020NEURIPS,DeVries2021ICCV,Chan2022CVPR,Schwarz2022NIPS}. They produce an intermediate 3D representation and render images for a given viewpoint with volume rendering~\cite{kajiya1984ray,Mildenhall2020ECCV}. 
Discriminator losses then ensure that the 3D representation  produces a valid image from any viewpoint. 
However, GANs come with notorious training instability and mode dropping behaviors~\cite{goodfellow2016nips,Arjovsky2017ICLR,li2018implicit}.
Denoising Diffusion models~\cite{Ho2020NIPS} (DDMs) have recently emerged as an alternative to GANs that avoid the aforementioned disadvantages
~\cite{salimans2022progressive,Rombach2022CVPR,saharia2022photorealistic}. 
However, DDMs model the data likelihood explicitly and are trained to reconstruct the training data. Thus, they have been used in limited scenarios~\cite{Zeng2022ARXIV,zheng2022neural} since ground-truth 3D data is not readily available at scale. 

To tackle the challenging problem of generating an entire scene with texture and geometry, we take inspiration from latent diffusion models (LDM)~\cite{Rombach2022CVPR}, which first construct an intermediate latent distribution of the training data then fit a diffusion model on the latent distribution.
In Sec.~\ref{sec:scene_ae}, we introduce a scene auto-encoder that encodes the set of RGB images into a neural field representation consisting of density and feature voxel grids.
To accurately capture a scene, the voxel grids' spatial dimension needs to be much larger than what current state-of-the-art LDMs can model.
In Sec.~\ref{sec:latent_ae}, we show how we can further compress and decompose the explicit voxel grids into compressed latent representations to facilitate learning the data distribution.
Finally, Sec.~\ref{sec:hldm} introduces a latent diffusion model that models the latent distributions in a hierarchical manner. Fig.~\ref{fig:pipeline} shows an overview of our method, which we name \ourmodel\ (\ourmodelsrt). 
We provide training and additional architecture details in the supplementary. 

\vspace{-1mm}
\subsection{Scene Auto-Encoder}
\label{sec:scene_ae}
\begin{figure}
  \centering
\includegraphics[width=0.95\linewidth]{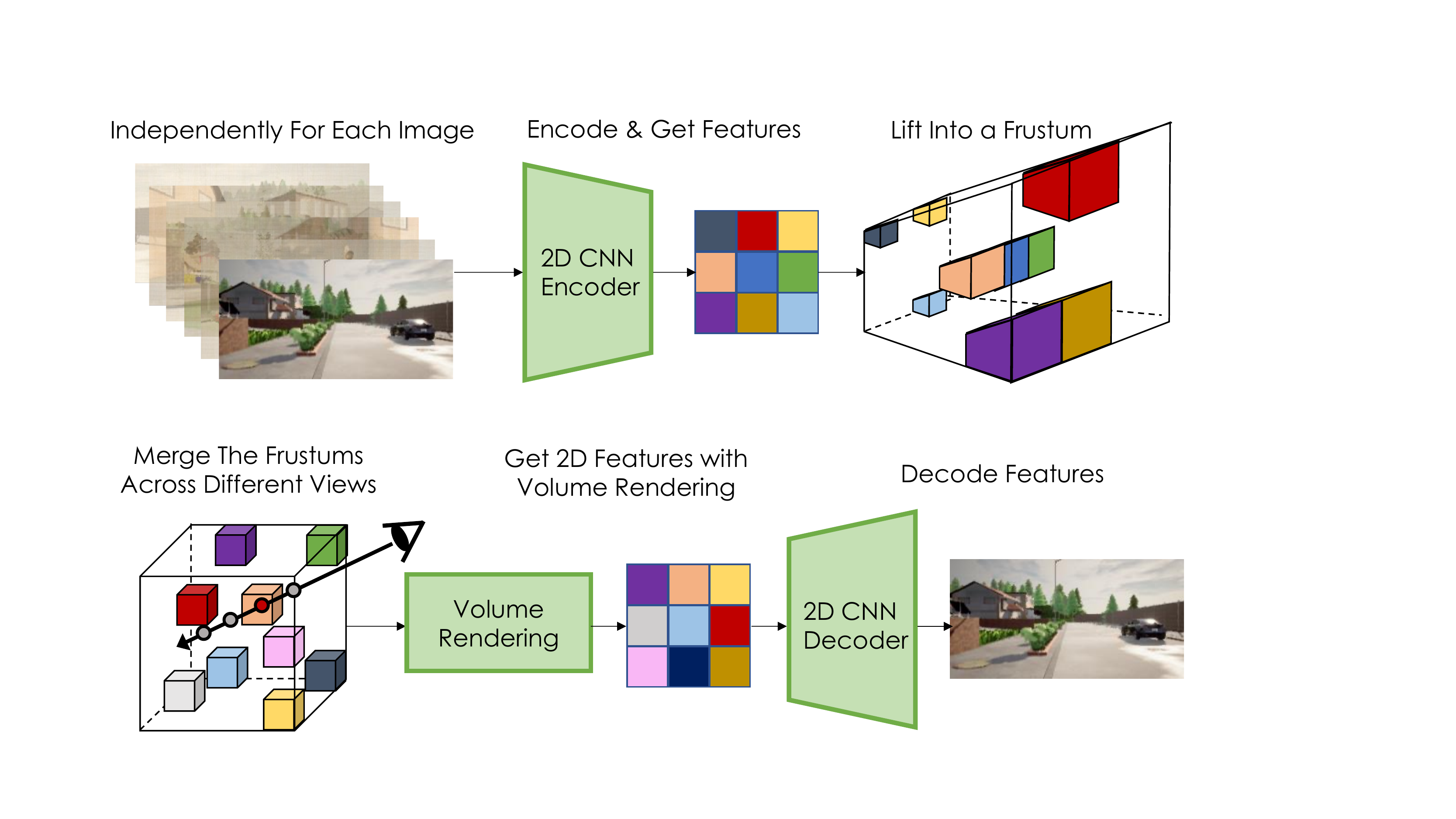}
\vspace{-3mm}
   \caption{
    \textbf{Scene Auto-Encoder:} Each input image is processed with a 2D CNN then lifted up to 3D and merged into the shared voxel grids. Density prediction is not shown here for brevity. 
   }
   \label{fig:scene_ae}
   \vspace{-3mm}
\end{figure}

The goal of the scene auto-encoder is to obtain a 3D representation of the scene from input images by learning to reconstruct them.
Fig.~\ref{fig:scene_ae} depicts the auto-encoding process.
The scene encoder is a 2D CNN and processes each RGB image $i_{1..N}$ separately, producing a $ \mathbb{R}^{H\times W \times (D+C)}$ dimensional 2D tensor for each image, where $H$ and $W$ are smaller than $i$'s size.   
We follow a similar procedure to Lift-Splat-Shoot (LSS)~\cite{Philion2020ECCV} to lift each 2D image feature map and combine them in the common voxel-based 3D neural field.
We build a discrete frustum of size $H\times W\times D$ with the camera poses $\kappa$ for each image.
This frustum contains image features and density values for each pixel, along a pre-defined discrete set of $D$ depths. 
Unlike LSS, we take the first $D$ channels of the 2D CNN's output and use them as density values. That is, the $d$'$th$ channel of the CNN's output at pixel $(h,w)$ becomes the density value of the frustum entry at $(h,w,d)$.
Motivated by the volume rendering equation~\cite{Mildenhall2020ECCV}, we get the occupancy weight $O$ of each element $(h, w, d)$ in the frustum using the density values $\sigma \ge 0$:
\begin{equation}
\label{eq:occupancy}
    O(h,w,d) = \exp(-\sum^{d-1}_{j=0}\sigma_{(h,w,j)}\delta_j)(1-\exp(-\sigma_{(h,w,d)}\delta_d))
\end{equation}
where $h,w$ denotes the pixel coordinate of the frustum and $\delta_j$ is the distance between each depth in the frustum.
Using the occupancy weights, we put the last $C$ channels of the CNN's output into the frustum $F$:
\begin{equation}
\label{eq:frustum}
F(h,w,d) = [O(h,w,d)\phi(h,w), \sigma(h,w,d)]
\end{equation}
where $\phi(h,w)$ denotes the $C$-channeled feature vector at pixel $(h,w)$ which is scaled by $O(h,w,d)$ for $F$ at depth $d$. 

After constructing the frustum for each view, we transform the frustums to world coordinates and fuse them into a shared 3D neural field, represented as density and feature voxel grids. 
Let $V_{\texttt{Density}}$ and $V_{\texttt{Feat}}$ denote the density and feature grid, respectively.
This formulation of representing a scene with density and feature grids has been explored before~\cite{Sun2021ARXIV} for optimization-based scene reconstruction and we utilize it as an intermediate representation for our scene auto-encoder. 
$V_{\texttt{Density,Feat}}$ have the same spatial size, and each voxel in $V$ represents a region in the world coordinate system.
For each voxel indexed by $(x,y,z)$, we pool all densities and features of the corresponding frustum entries.
In this paper, we simply take the mean of the pooled features.  More sophisticated pooling functions (\eg. attention) can be used, which we leave as future work. 

Finally, we perform volume rendering using the camera poses $\kappa$ to project $V$ onto a 2D feature map. 
We trilinearly interpolate the values on each voxel to get the feature and density for each sampling point along the camera rays. 2D features are then fed into a CNN decoder that produces the output image $\hat{i}$. 
We denote rendering of voxels to output images as $\hat{i} = r(V,\kappa)$.
From the volume rendering process, we also get the expected depth $\hat{\rho}$ along each ray~\cite{Rematas2022CVPR}.
The scene auto-encoding pipeline is trained with an image reconstruction loss $||i-\hat{i}||$ and a depth supervision loss $||\rho-\hat{\rho} ||$.
In the case of sparse depth measurements, we only supervise the pixels with recorded depth. 
We can further improve the quality of the auto-encoder with adversarial loss as in VQGAN~\cite{Esser2021CVPR} or by doing a few optimization steps at inference time, which we discuss in the supplementary.

\subsection{Latent Voxel Auto-Encoder}
\vspace{-1mm}
\label{sec:latent_ae}
It is possible to fit a generative model on voxel grids obtained from Sec.~\ref{sec:scene_ae}.
However, to capture real-world scenes, the dimensionality of the representation needs to be much larger than what SOTA diffusion models can be trained on. For example, Imagen~\cite{saharia2022photorealistic} trains DDMs on $256\times 256$ RGB images, and we use voxels of size $128\times 128\times 32$ with $32$ channels. 
We thus introduce a latent auto-encoder (LAE) that 
compresses voxels into a $128$-dimensional global latent as well as coarse (3D) and fine (2D) quantized latents with channel dimensions of four and spatial dimensions $32\times 32\times 16$ and $128\times 128$ respectively. 

We concatenate $V_{\texttt{Density}}$ and $V_{\texttt{Feat}}$ along the channel dimension and use separate CNN encoders to encode the voxel grid $V$ into a hierarchy of three latents: 1D global latent $g$, 3D coarse latent $c$, and 2D fine latent $f$, as shown in Fig.~\ref{fig:pipeline}.
The intuition for this design is that $g$ is responsible for representing the global properties of the scene, such as the time of the day, $c$ represents coarse 3D scene structure, and $f$ is a 2D tensor with the same horizontal size $X\times Y$ as $V$, which gives further details for each location $(x,y)$ in bird's eye view perspective.  
We empirically found that 2D CNNs perform similarly to 3D CNNs while being more efficient, thus we use 2D CNNs throughout.
To use 2D CNNs for the 3D input $V$, we concatenate $V$'s vertical axis along the channel dimension and feed it to the encoders. 
We also add latent regularizations to avoid high variance latent spaces~\cite{Rombach2022CVPR}. 
For the 1D vector $g$, we use a small KL-penalty via the reparameterization trick~\cite{Kingma2014ICLR}, and for $c$ and $f$, we impose a vector-quantization~\cite{van2017neural,Esser2021CVPR} layer to regularize them.

The CNN decoder is similarly a 2D CNN, and takes $c$, concatenated along vertical axis, as the initial input. The decoder uses conditional group normalization layers~\cite{wu2018group} with $g$ as the conditioning variable.
Lastly, we concatenate $f$ to an intermediate tensor in the decoder. 
The latent decoder outputs $\hat{V}$ which is the reconstructed voxel. 
LAE is trained with the voxel reconstruction loss $||V-\hat{V}||$ along with the image reconstruction loss $||i-\hat{i}||$ where $\hat{i} = r(\hat{V},\kappa)$. 
Note that the image reconstruction loss only helps with the learning of LAE, and the scene auto-encoder is kept fixed.

\vspace{-0mm}
\subsection{Hierarchical Latent Diffusion Models}
\label{sec:hldm}

\begin{figure}[t!]
  \centering
\includegraphics[width=0.9\linewidth]{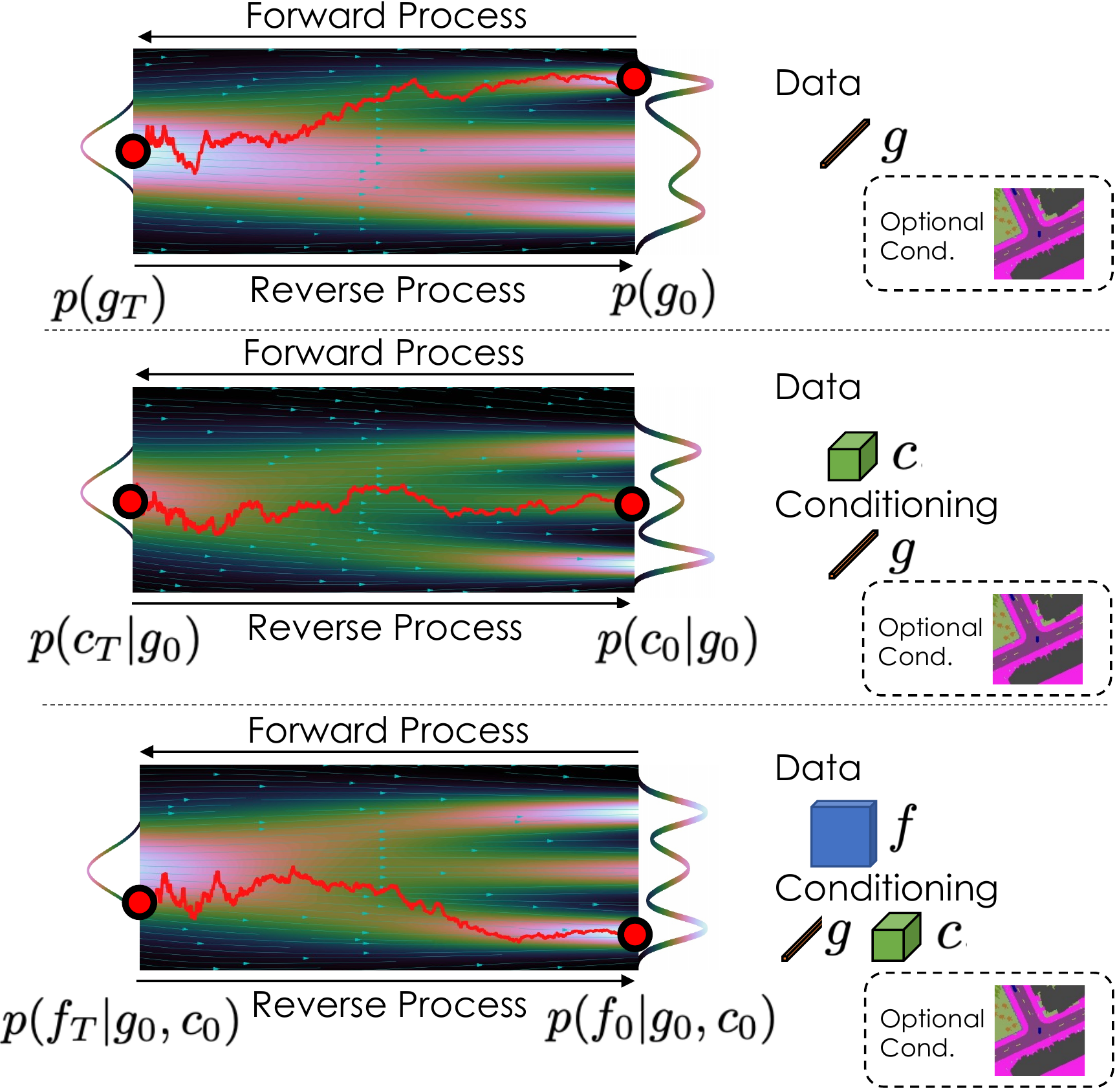}
\vspace{-2mm}
   \caption{
   \textbf{Hierarchical LDM.}
   \textbf{Top:} LDM $\psi_g$ for KL-regulairzed global latent $g$.
   \textbf{Middle:} LDM $\psi_c$ for vector-quantized coarse latent $c$.
   \textbf{Bottom:} LDM $\psi_f$ for vector-quantized fine latent $f$.
    All LDMs optionally take an additional conditioning variable as input, such as a Bird's Eye View segmentation map as depicted here.
   }
   \label{fig:hldm}
   \vspace{-3.5mm}
\end{figure}

Given the latent variables $g, c, f$ that represent a voxel-based scene representation $V$, we define our generative model as $p(V,g,c,f) = p(V|g,c,f)p(f|g,c)p(c|g)p(g)$ with Denoising Diffusion Models 
(DDMs)~\cite{ho2020ddpm}. 
In general, DDMs with discrete time steps have a fixed Markovian forward process $q(x_t|x_{t-1})$ where $q(x_0)$ denotes the data distribution and $q(x_T)$ is defined to be close to the standard normal distribution, where we use the subscript to denote the time step.
DDMs then learn to revert the forward process $p_{\theta}(x_{t-1}|x_t)$ with learnable parameters $\theta$. 
It can be shown that learning the reverse process is equivalent to learning to denoise $x_t$ to $x_0$ for all timesteps $t$~\cite{ho2020ddpm,ho2022video} by reducing the following loss:\\[-3mm]
\begin{equation}
\mathop{\mathbb{E}}_{t,\epsilon,x_0} \Bigl[ w(\lambda_t)||x_0-\hat{x}_{\theta}(x_t, t)||^2_2 \Bigr]
\end{equation}
where $t$ is sampled from a uniform distribution for timesteps, $\epsilon$ is sampled from the standard normal to noise the data $x_0$, $w(\lambda_t)$ is a timestep dependent weighting constant, and $\hat{x}_{\theta}$ denotes the learned denoising model.

We train our hierarchical LDM with the following losses:\\[-4mm]
\begin{align}
&\mathop{\mathbb{E}}_{t,\epsilon,g_0} \Bigl[ w(\lambda_t)||g_0-\psi_g(g_t, t)||^2_2 \Bigr] \\
&\mathop{\mathbb{E}}_{t,\epsilon,g_0,c_0} \Bigl[ w(\lambda_t)||c_0-\psi_c(c_t,g_0, t)||^2_2 \Bigr] \\
&\mathop{\mathbb{E}}_{t,\epsilon,g_0,c_0,f_0} \Bigl[ w(\lambda_t)||f_0-\psi_f(f_t,g_0,c_0, t)||^2_2 \Bigr]
\end{align}
where $\psi$ denotes the learnable denoising networks for $g,c,f$. 
Fig.~\ref{fig:hldm} visualizes the diffusion models. 
$\psi_g$ is implemented with linear layers with skip connections and $\psi_c$ and $\psi_f$ adopt the U-net architecture~\cite{ronneberger2015u}. $g$ is fed into $\psi_c$ and $\psi_f$ with conditional group normalization layers. $c$ is interpolated and concatenated to the input to $\psi_f$. 
The camera poses contain the trajectory the camera is travelling, and this information can be useful for modelling a 3D scene as it tells the model where to focus on generating. 
Therefore, we concatenate the camera trajectory information to $g$ and also learn to sample it. For brevity, we still call the concatenated vector $g$. 
For conditional generation, each $\psi$ takes the conditioning variable as input with cross-attention layers~\cite{Rombach2022CVPR}.

Each $\psi$ can be trained in parallel and, once trained, can be sampled one after another following the hierarchy. 
In practice, we use the v-prediction parameterization~\cite{salimans2022progressive} that has been shown to have better convergence and training stability~\cite{salimans2022progressive,saharia2022photorealistic}.
Once $g,c,f$ are sampled, we can use the latent decoder from Sec.~\ref{sec:latent_ae} to construct the voxel $V$ which represents the neural field for the sampled scene.. 
Following the volume rendering and decoding step in Sec.~\ref{sec:scene_ae}, the sampled scene can be visualized from desired viewpoints.

\vspace{-1mm}
\subsection{Post-Optimizing Generated Neural Fields}
\label{sec:method_post_opt}

Samples generated from our model on real-world data contain reasonable texture and geometry (Fig.~\ref{fig:post_opt}), but can be further optimized 
by leveraging recent advances in 2D image diffusion models trained on orders of magnitude more data.
Specifically, we iteratively update initially generated voxels, $V$, by rendering viewpoints from the scene and applying Score Distillation Sampling (SDS)~\cite{poole2022dreamfusion} loss on each image independently:
\begin{equation}
\label{eq:sds_grad}
\nabla_V L_{SDS} = \mathop{\mathbb{E}}_{\epsilon,t,\kappa} \Bigl[ w(\lambda_t)(\epsilon-\hat{\epsilon_{\theta}}(r(V, \kappa), t ) ) \frac{\partial r(V, \kappa)}{\partial V} \Bigr]
\end{equation}
where $\kappa$ is sampled uniformly in a $6m^2$ region around the origin of the scene with random rotation about the vertical axis, $w(\lambda_t)$ is the weighting schedule used to train $\hat{\epsilon_{\theta}}$ and $t \sim U[0.02T,0.2T]$ where $T$ is the amount of noise steps used to train $\hat{\epsilon_{\theta}}$.
Note that for latent diffusion models, the noise prediction step is applied after encoding $r(V, \kappa)$ to the LDM's latent space and the partial gradient term is updated appropriately.
For $\hat{\epsilon_{\theta}}$, we use an off-the-shelf latent diffusion model~\cite{Rombach2022CVPR}, finetuned to condition on CLIP image embeddings~\cite{Radford2021ARXIV}\footnote[1]{https://github.com/justinpinkney/stable-diffusion}.
We found that CLIP contains a representation of the quality of images that the LDM is able to interpret: denoising an image while conditioning on CLIP image embeddings of our model's samples produced images with similar geometry distortions and texture errors.
We leverage this property by optimizing $L_{SDS}$ with negative guidance.
Letting $y,y'$ be CLIP embeddings of clean image conditioning (\eg dataset images) and artifact conditioning (\eg samples) respectively, we perform classifier-free guidance~\cite{ho2022classifier} with conditioning vector $y$, but replace the unconditional embedding with $y'$. 
As shown in the supplementary, this is equivalent (up to scale) to sampling from $\frac{p(x|y)^{\alpha}}{p(x|y')}$ at each denoising step where $\alpha$ controls the trade-off between sampling towards dataset images and away from images with artifacts.
We stress that this post-optimization is only successful due to the strong scene prior contained in our voxel samples, as shown by our comparison to running optimization on randomly initialized voxels in the supplementary. 


%% file: sections/experiments.tex
\begin{figure}[t!]
  \centering
\includegraphics[width=1.0\linewidth]{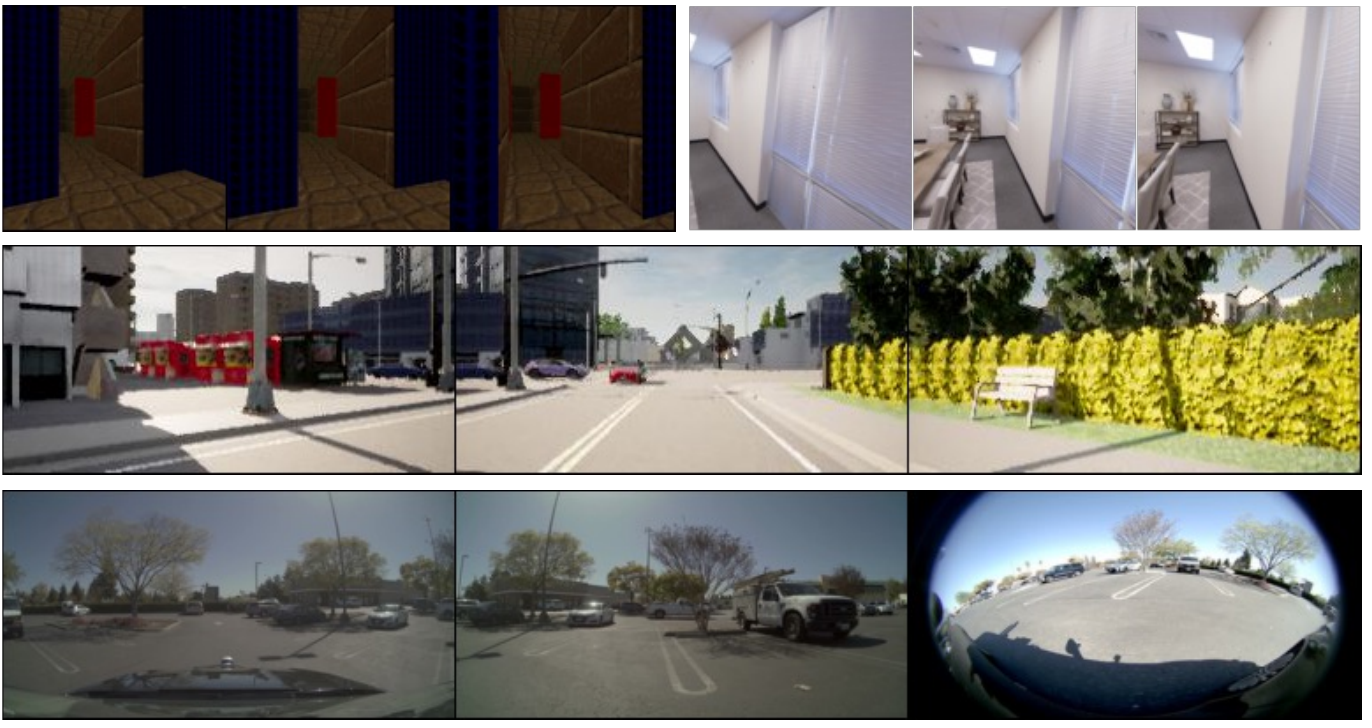}
\vspace{-6mm}
   \caption{{\bf Datasets:} Top-left: VizDoom~\cite{Keuper2018PAMI}. Top-right: Replica~\cite{Straub2019ARXIV}. Middle: Carla~\cite{Dosovitskiy2017CORL}
   \label{fig:dataset}. Bottom: AVD. For Carla and AVD, we visualize a subset of available cameras. }
     \vspace{-2mm}
\end{figure}

\begin{figure}[t!]
  \centering
\includegraphics[width=1.0\linewidth]{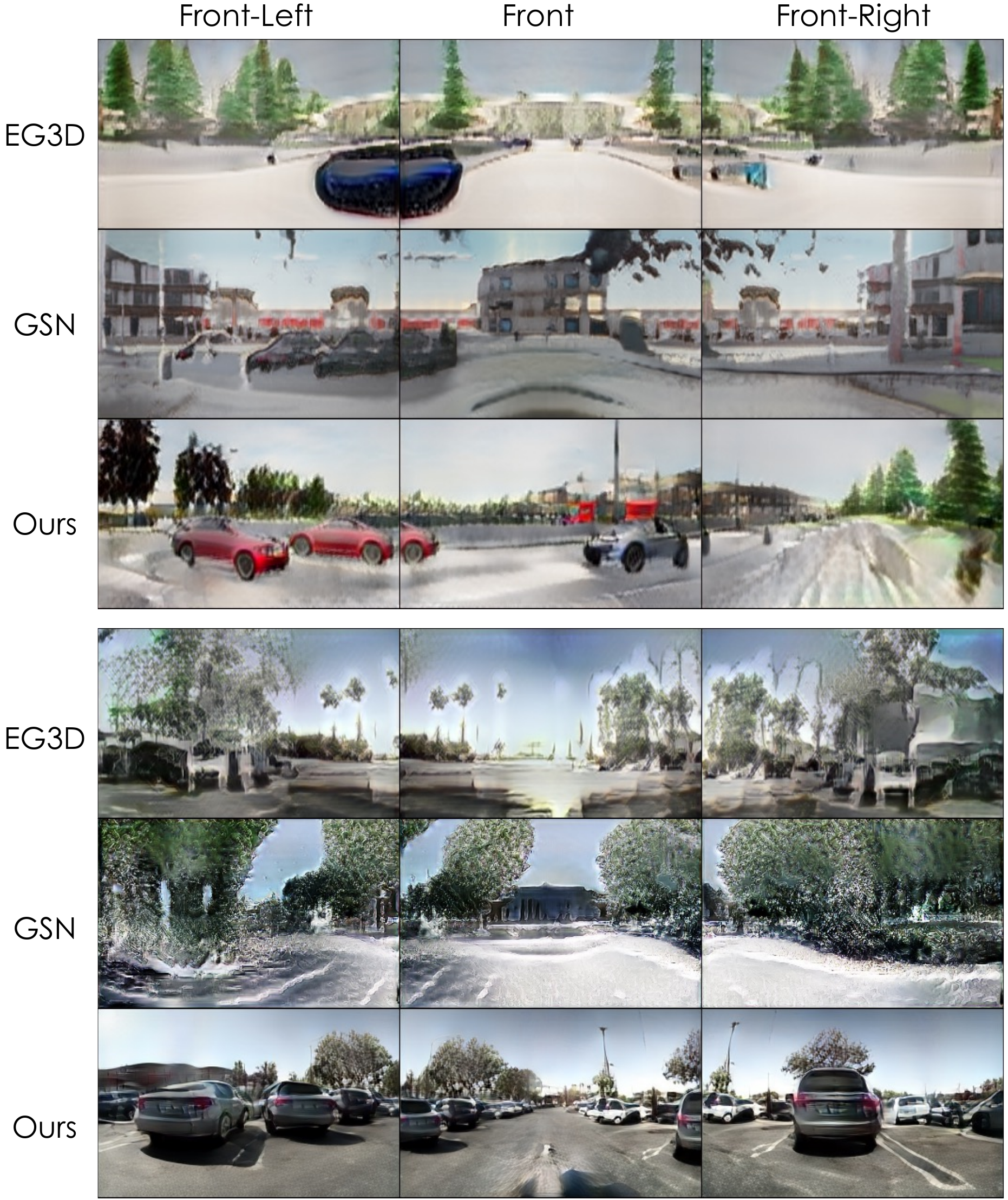}
\vspace{-6mm}
   \caption{\textbf{Generated Scenes:} The top three rows are samples from Carla, and the bottom three rows are samples from AVD. }
   \label{fig:sample_comparison}
   \vspace{-3mm}
\end{figure}

\begin{figure}[t!]
  \centering
\includegraphics[width=0.8\linewidth]{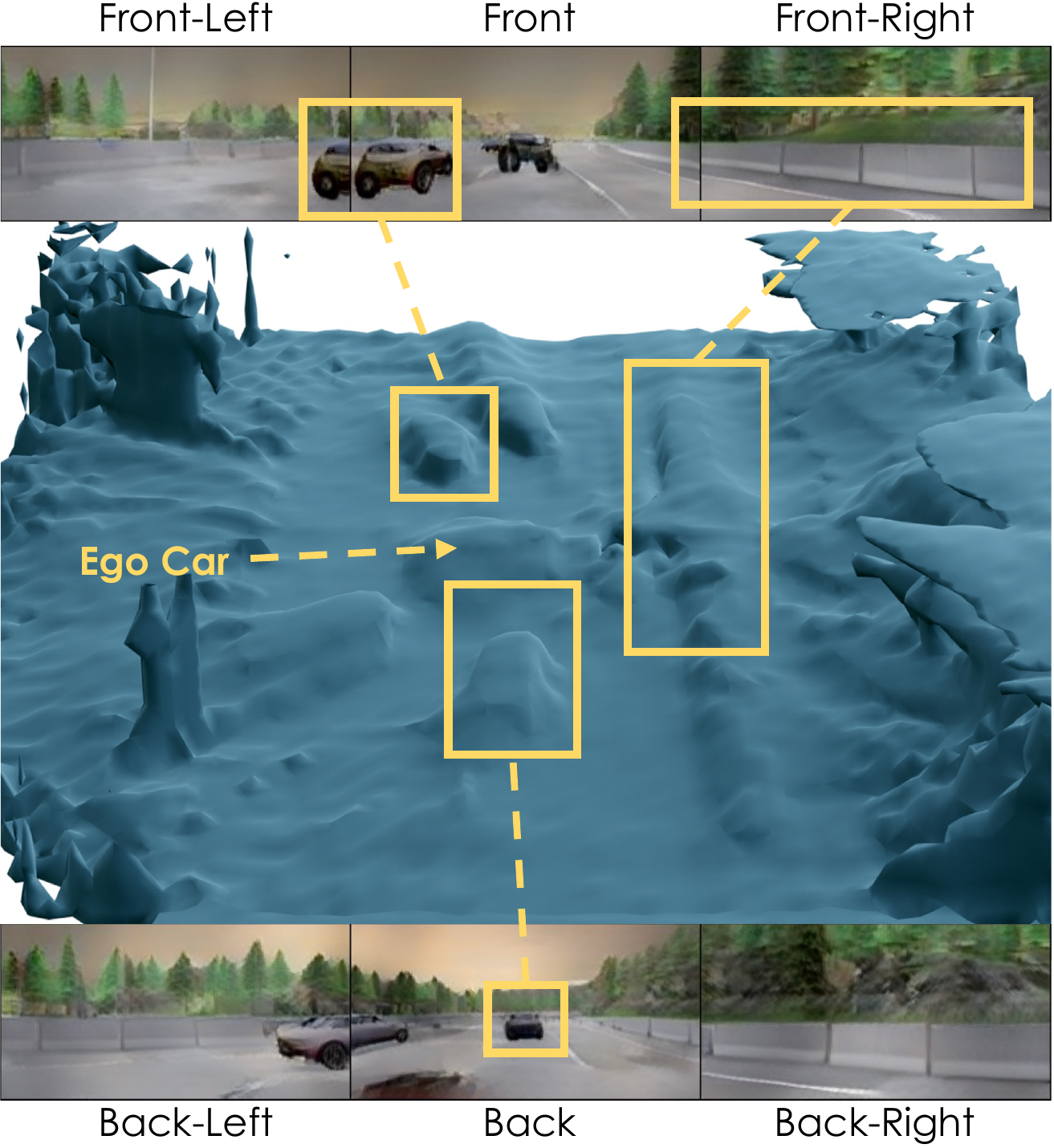}
\vspace{-3mm}
   \caption{We run marching-cubes~\cite{Lorensen1987SIGGRAPH} on the density voxels to visualize the geometry of the samples generated by \ourmodelsrt. The structure of the scene is reflected well in the mesh.}
   \label{fig:mesh_vis}
   \vspace{-6mm}
\end{figure}

\section{Experiments}
\label{sec:experiments}

We evaluate \ourmodel\ on the following four datasets (Fig.~\ref{fig:dataset}). Each dataset contains RGB images and a depth measurement with their corresponding camera poses.

\textbf{VizDoom}~\cite{kempka2016vizdoom} consists of front-view sensor observations obtained by navigating inside a synthetic game environment.
We use the dataset provided by~\cite{devries2021unconstrained}, which contains 34 trajectories with a length of 600. 

\textbf{Replica}~\cite{Straub2019ARXIV} is a dataset of high-quality reconstructions of 18 indoor scenes, containing 101 front-view trajectories with a length of 100. 
We use the dataset provided by \cite{devries2021unconstrained}.

\textbf{Carla}~\cite{Dosovitskiy2017CORL} is an open-source simulation platform for self-driving research. 
We mimic the camera settings for a self-driving car, by placing six cameras \textit{(front-left, front, front-right, back-left, back, back-right)}, covering 360 degrees with some overlaps, and move the car in a randomly sampled direction and distance for 10 timesteps.
We sample 43K datapoints, each containing 60 images.

\textbf{AVD} is an in-house dataset of human driving recordings in roads and parking lots. 
It has ten cameras with varying lens types along with Lidar for depth measurement. 
It has 73K sequences, each with 8 frames extracted at 10 fps.

\subsection{Baseline Comparisons}

\input{sections/tables/fid_VD_Replica.tex} 
\input{sections/tables/fid_CarlaAv.tex}

\begin{figure}[t!]
\vspace{-2mm}
  \centering
\includegraphics[width=0.9\linewidth]{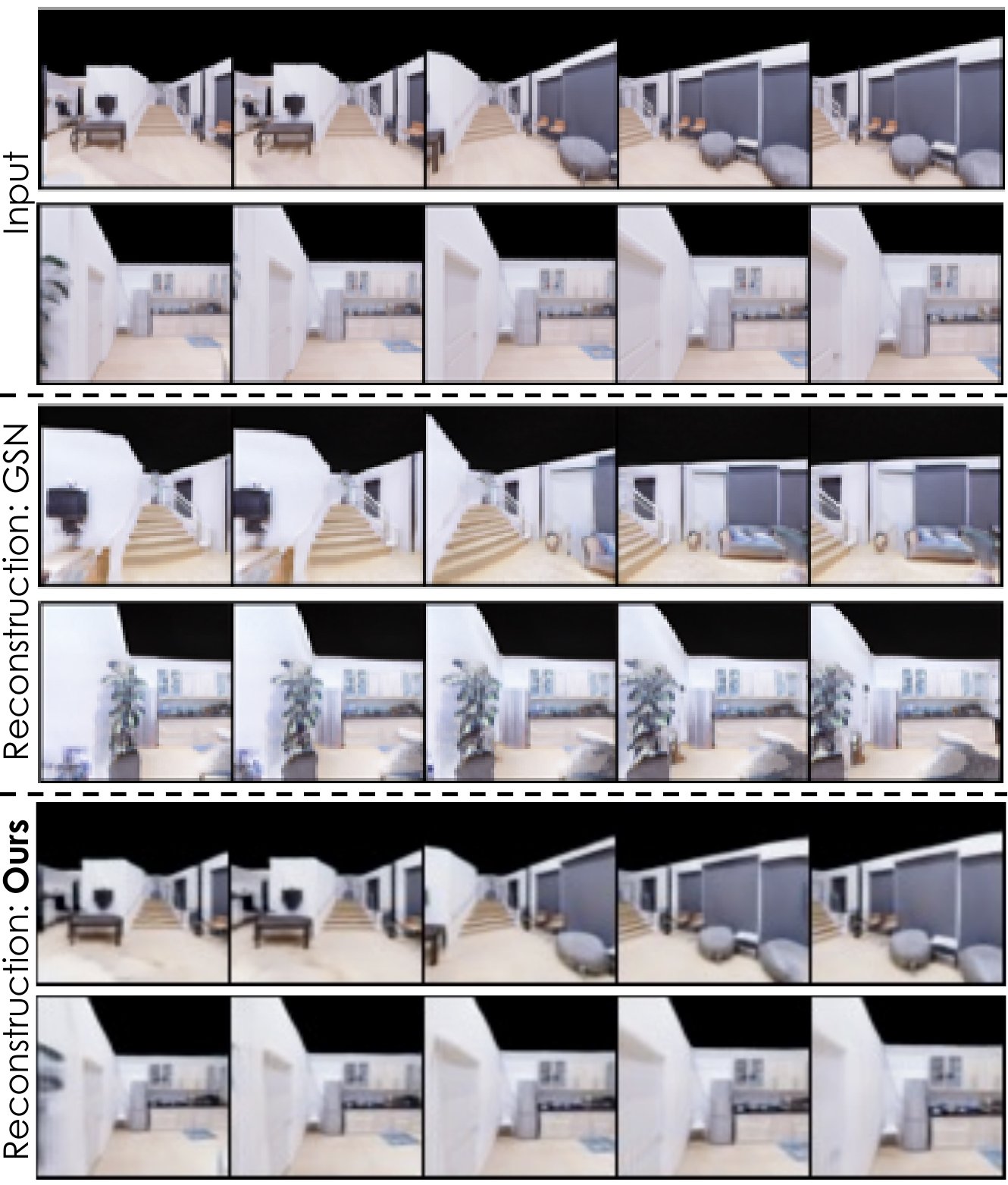}
\vspace{-2mm}
   \caption{Reconstructing held-out scenes not seen during training.}
   \label{fig:rep_recon}
   \vspace{-4mm}
\end{figure}

\textbf{Unconditional Generation}
We evaluate the unconditional generation performance of \ourmodelsrt\ by comparing it with baseline models.
All results are without the post-optimization step (Sec.~\ref{sec:method_post_opt}), unless specified.
Tab.~\ref{tab:fid_vd_rep} shows the results on VizDoom and Replica.
GRAF~\cite{Schwarz2020NEURIPS} and $\pi$-GAN~\cite{Chan2020ARXIV}, which do not utilize ground truth depth in training, have shown successes in modelling single objects, but they exhibit worse performance than others that leverages depth information for modelling scenes. 
GAUDI~\cite{Bautista2022ARXIV} is an auto-decoder-based diffusion model.
Their auto-decoder optimizes a small per-scene latent to reconstruct its matching scene. 
GAUDI comes with the advantage that learning the generative model is simple as it only needs to model the small dimensional latent distribution that acts as the key to their corresponding scenes.
On the contrary, {\ourmodelsrt} is trained on the latents that are a decomposition of the explicit 3D neural field.
Therefore, GAUDI puts more modelling capacity into the auto-decoder part, and {\ourmodelsrt} puts it more into the generative model part. 
We attribute our improvement over GAUDI to our expressive hierarchical LDM that can model the details of the scenes better. 
In VizDoom, only one scene exists, and each sequence contains several hundred steps covering a large area in the scene, which our voxels were not large enough to encode. 
Therefore, we chunked each VizDoom trajectory to be 50 steps long.

Tab.~\ref{tab:fid_carla_avd} shows results on complex outdoor datasets: Carla and AVD. 
We compare with EG3D~\cite{Chan2021eg3d} and GSN~\cite{DeVries2021ICCV}.
Both are GAN-based 3D generative models, but GSN utilizes ground truth depth measurements.
Note that we did not include GAUDI~\cite{Bautista2022ARXIV} as the code was not available.
\ourmodelsrt\ achieves the best performance, and both baseline models have difficulty modelling the real outdoor dataset (AVD).  
Fig.~\ref{fig:sample_comparison} compares the samples from different models. 

Since the datasets are composed of frame sequences, we can treat them as videos and further evaluate with Fr\'echet Video Distance (FVD)~\cite{unterthiner2018towards} to compare the distributions of the dataset and sampled sequences. 
This can quantify samples' 3D structure by how natural the rendered sequence from a moving camera is.
For EG3D and GSN, we randomly sample a trajectory from the datasets and for \ourmodelsrt, we sample a trajectory from the global latent diffusion model. 
Tab.~\ref{tab:fvd_carla_avd} shows that {\ourmodelsrt} achieves the best results. 
We empirically observed GSN sometimes produced slightly inconsistent rendering, which could attribute to its lower FVD score than EG3D's.
We also visualize the geometry of \ourmodelsrt's samples by running marching-cubes~\cite{Lorensen1987SIGGRAPH} on the density voxels.
Fig.~\ref{fig:mesh_vis} shows that our samples produce a coarse but realistic geometry. 


\input{sections/tables/fvd_CarlaAv.tex}

\begin{figure}
  \centering
\includegraphics[width=1.0\linewidth]{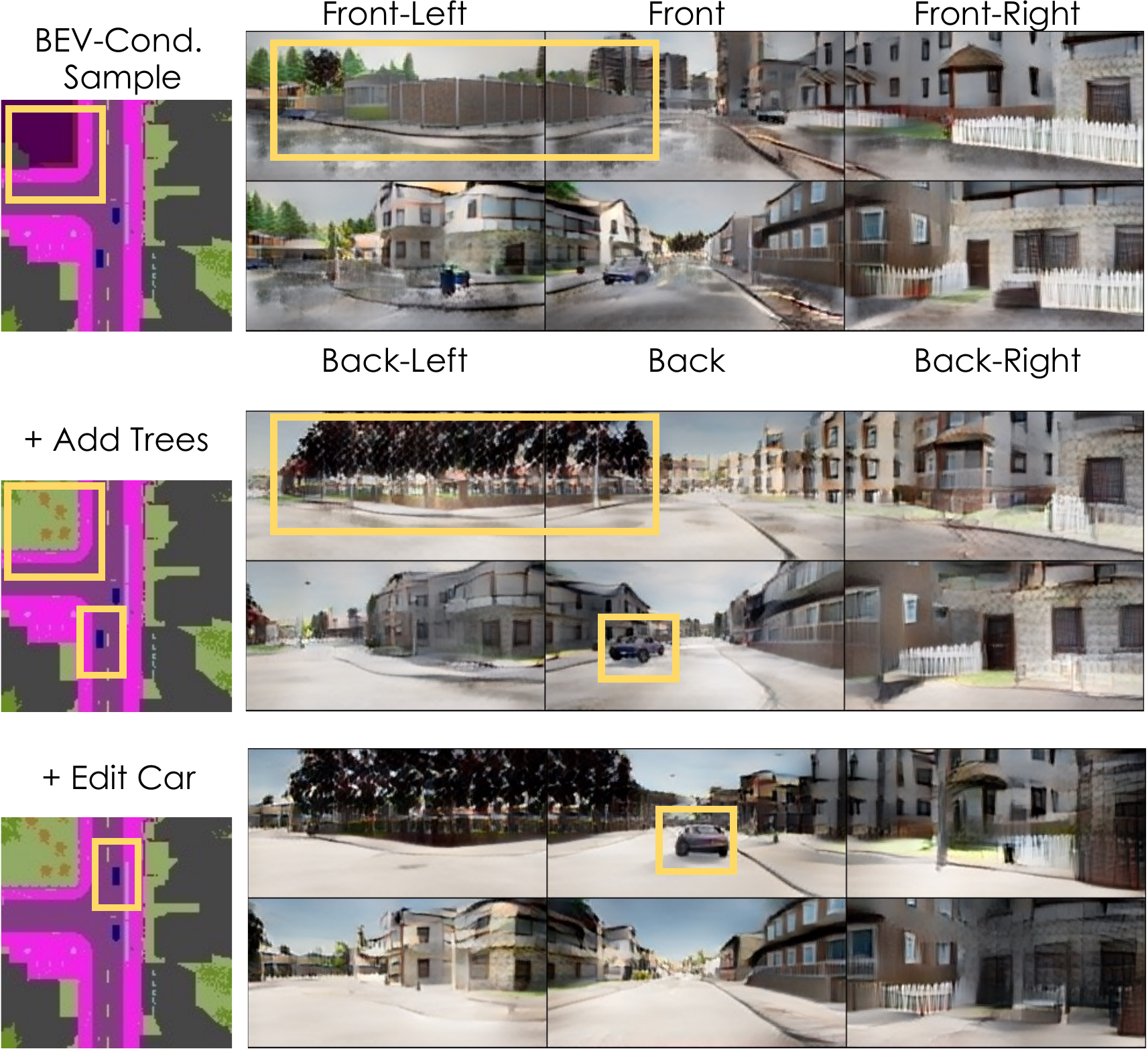}
\vspace{-6mm}
   \caption{{\bf BEV-Conditioned Synthesis}: \ourmodelsrt\ allows controllable generation by editing the BEV segmentation map. From the initial sample, we add trees (green) and then edit the location of the car (blue). Note the ego car is at the center and thus not rendered.}
   \label{fig:bev}
   \vspace{-3mm}
\end{figure}

\begin{figure*}[!t]
  \centering
\includegraphics[width=0.85\textwidth]{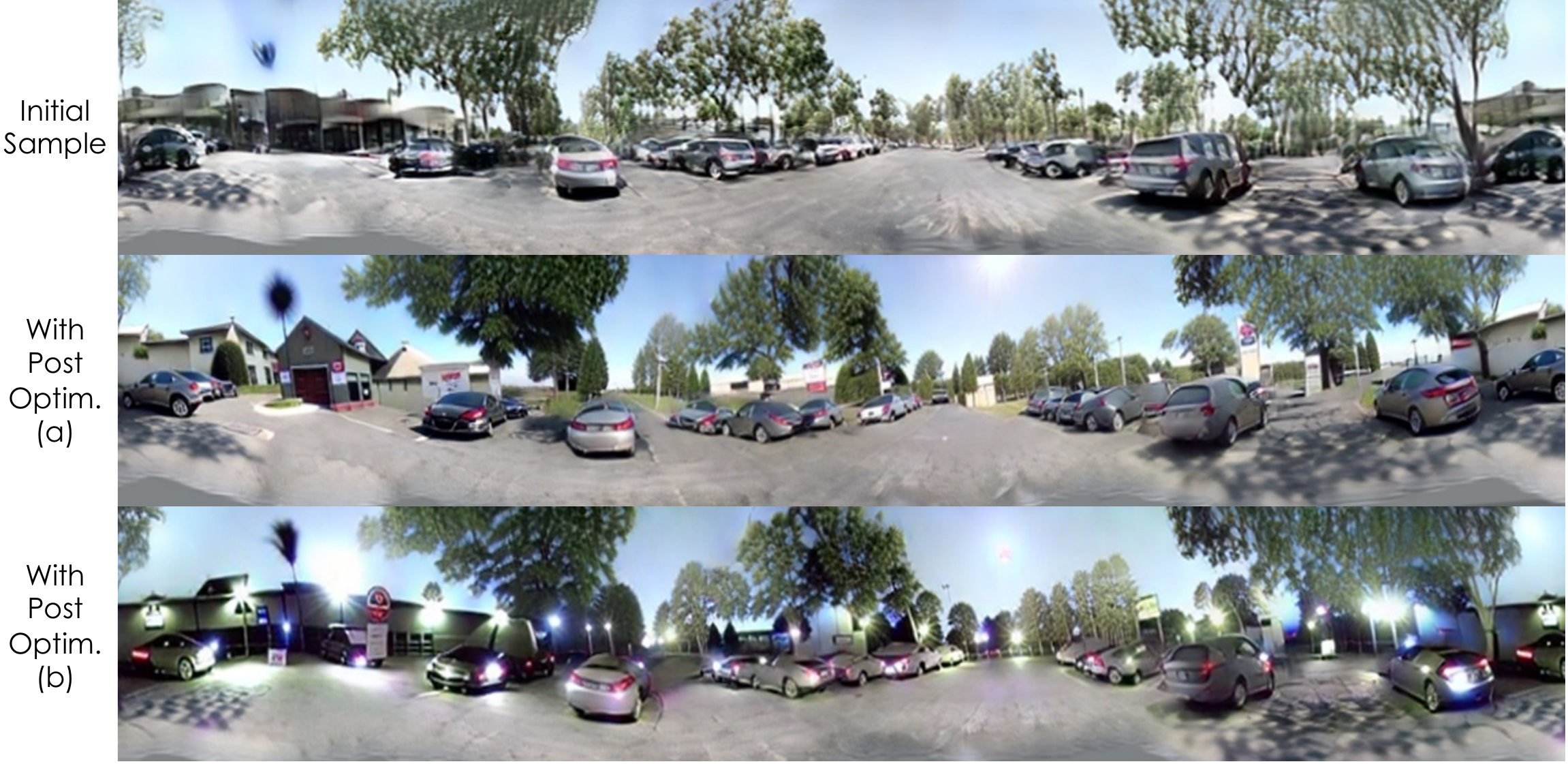}
   \vspace{-3mm}
   \caption{ 
   \textbf{Panoramas from {\ourmodelsrt}'s samples:} From the initial sample at the top, we apply post-optimization with Score Distillation Sampling~\cite{poole2022dreamfusion} (Sec.~\ref{sec:method_post_opt}). (a) demonstrates improved sample quality. (b) showcases style modification by conditioning on evening scenes.
   }
   \vspace{-3mm}
\label{fig:post_opt}
\end{figure*}

\textbf{Ablations} 
We evaluate the hierarchical structure of \ourmodelsrt. 
Tab.~\ref{tab:ablation} shows an ablation study on Carla.
The model with the full hierarchy achieves the best performance. 
The global latent makes it easier for the other LDMs to sample as conditioning on the global properties of the scene (\eg time of day) narrows down the distribution they need to model.
The 2D fine latent helps retain the residual information missing in the 3D coarse latent, thus improving the latent auto-encoder and, consequently, the LDMs.

\input{sections/tables/ablation.tex}

\begin{figure}
  \centering
\includegraphics[width=1.0\linewidth]{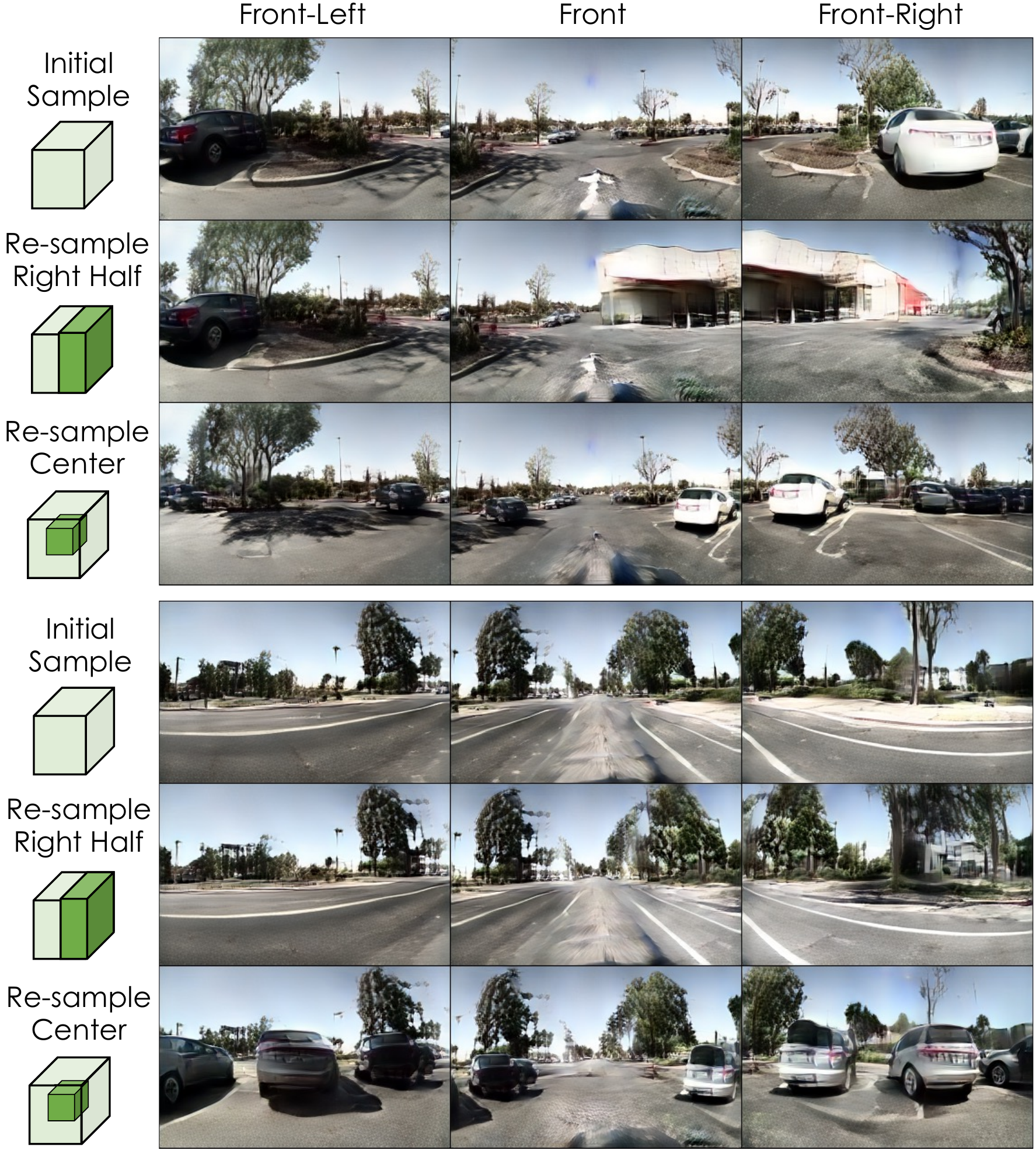}
\vspace{-6mm}
   \caption{{\bf Scene Editing}: We use the 3D coarse latent $c$ for scene editing. From the initial sample indicated by light green, we re-sample a part of the latent, indicated by dark green. }
   \label{fig:av_editing}
     \vspace{-2mm}
\end{figure}

\textbf{Scene Reconstruction}
Unlike previous approaches, {\ourmodelsrt} has an explicit scene auto-encoder that can be used for scene reconstruction. 
GAUDI~\cite{Bautista2022ARXIV} is auto-decoder based, so it is not trivial to infer a latent for a new scene. 
GSN~\cite{DeVries2021ICCV} can invert a new scene using a GAN inversion method~\cite{richardson2021encoding,zhu2020domain}, but as Fig.~\ref{fig:rep_recon} shows, it fails to get the details of the scene correct. 
Our scene auto-encoder generalizes well and is scalable as the number of scenes grow.

\subsection{Applications and Limitations}
\label{sec:applications}

\textbf{Conditional Synthesis} \ourmodelsrt\ can utilize additional conditioning signals for controllable generation. 
In this paper, we consider Bird's Eye View (BEV) segmentation maps, but our model can be extended to other conditioning variables. 
We use cross attention layers~\cite{Rombach2022CVPR}, which have been shown to be effective for conditional synthesis. 
Fig.~\ref{fig:bev} shows that \ourmodelsrt\ follows the given BEV map faithfully and how the map can be edited for controllable synthesis.

\textbf{Scene Editing} 
Image diffusion models can be used for image inpainting without explicit training on the task~\cite{song2020score,Rombach2022CVPR}.
We leverage this property to edit scenes in 3D by re-sampling a region in the 3D coarse latent $c$.
Specifically, at each denoising step, we noise the region to be kept and concatenate with the region being sampled, and pass it through the diffusion model. 
We use reconstruction guidance~\cite{ho2022video} to better harmonize the sampled and kept regions.
After we get a new $c$, the fine latent is also re-sampled conditioned on $c$.
Fig.~\ref{fig:av_editing} shows results on scene editing with \ourmodelsrt.

\textbf{Post-Optimization} 
Fig.~\ref{fig:post_opt} shows how post-optimization (Sec.~\ref{sec:method_post_opt}) can improve the quality of {\ourmodelsrt}'s initial sample while retaining the 3D structure. 
In addition to improving quality, we can also modify scene properties, such as time of day and weather, by conditioning the LDM on images with the desired properties. 
SDS-based style modification is effective for changes where a set of clean image data is available with the desired property and is reasonably close to our dataset's domain (\eg street images for AVD). 
In the supplementary, we also provide results experimenting with directional CLIP loss~\cite{gal2021stylegannada} to quickly finetune our scene decoder for a given text prompt.


\textbf{Limitations}
\ourmodelsrt's hierarchical structure and three stage pipeline allows us to achieve high-quality generations and reconstructions, but it comes with a degradation in training time and sampling speed. 
In this work, the neural field representation is based on dense voxel grids, and it becomes expensive to volume render and learn the diffusion models as they get larger.
Therefore, exploring alternative sparse representations is a promising future direction.
Lastly, our method requires multi-view images which limits data availability and therefore risks universal problems in generative modelling of overfitting.
For example, we found that output samples in AVD had limited diversity because the dataset itself was recorded in a limited number of scenes. 

%% file: sections/tables/fid_VD_Replica.tex
\begin{table}
\centering
  \resizebox{0.9\linewidth }{!}{
\begin{tabular}{l|c|c|cc}
    \toprule
    Criterion & Method  & Depth & VizDoom & Replica \\
    \midrule
   \multirow{5}{*}{FID ($\downarrow$)}
    & GRAF~\cite{Schwarz2020NEURIPS} & \xmark & 47.50 & 65.37  \\
    & $\pi$-GAN~\cite{Chan2020ARXIV} & \xmark & 143.55 & 166.55  \\
    & GSN~\cite{DeVries2021ICCV} & \cmark & 37.21 & 41.75  \\
    & GAUDI~\cite{Bautista2022ARXIV} & \cmark & 33.70 & 18.75  \\
    & \ourmodelsrt & \cmark & $\textbf{19.54}^{\ast}$ & \textbf{14.59}  \\
    \bottomrule
\end{tabular}
}
\vspace{-3mm}
\caption{
FID~\cite{Heusel2017NIPS} scores on VizDoom and Replica. \ourmodelsrt\ outperforms all baseline models. Baseline numbers are from \cite{Bautista2022ARXIV}.
}
\vspace{-2mm}
\label{tab:fid_vd_rep}
\end{table}

%% file: sections/tables/fid_CarlaAv.tex
\begin{table}
\centering
  \resizebox{0.85\linewidth }{!}{
\begin{tabular}{l|c|c|cc}
    \toprule
    Criterion & Method  & Depth & Carla & AVD \\
    \midrule
   \multirow{3}{*}{FID ($\downarrow$)}
    & EG3D~\cite{Chan2021eg3d} & \xmark & 76.89 & 194.34  \\
    & GSN~\cite{DeVries2021ICCV} & \cmark & 75.45 & 166.07  \\
    & \ourmodelsrt & \cmark & \textbf{35.69} & \textbf{54.26}  \\
    \bottomrule
\end{tabular}
}
\vspace{-2mm}
\caption{
FID~\cite{Heusel2017NIPS} scores on Carla and AVD datasets. Baseline models have trouble learning the distribution of complex outdoor datasets, in particular AVD, while \ourmodelsrt\ models them well.
}
\vspace{-2mm}
\label{tab:fid_carla_avd}
\end{table}

%% file: sections/tables/fvd_CarlaAV.tex
\begin{table}
\centering
  \resizebox{0.85\linewidth }{!}{
\begin{tabular}{l|c|c|cc}
    \toprule
    Criterion & Method  & Depth & Carla & AVD \\
    \midrule
   \multirow{3}{*}{FVD ($\downarrow$)}
    & EG3D~\cite{Chan2021eg3d} & \xmark & 134.94 & 1232.38  \\
    & GSN~\cite{DeVries2021ICCV} & \cmark & 184.30 & 1659.81  \\
    & \ourmodelsrt & \cmark & \textbf{91.80} & \textbf{242.50}  \\
    \bottomrule
\end{tabular}
}
\vspace{-2mm}
\caption{
FVD~\cite{unterthiner2018towards} scores on Carla and AVD Datasets. As for FID, baseline models have trouble learning to model complex datasets.
}
\vspace{-2mm}
\label{tab:fvd_carla_avd}
\end{table}

%% file: sections/tables/ablation.tex

\begin{table}
\centering
  \resizebox{0.85\linewidth }{!}{
\begin{tabular}{c|ccc}
    \toprule
     & Coarse lat. $c$  &  + Fine lat. $f$ & + Global  lat. $g$  \\
    \midrule
    FID ($\downarrow$) & 46.43 & 43.52 & \textbf{35.69}  \\
    \bottomrule
\end{tabular}
}
\vspace{-3mm}
\caption{
FID~\cite{Heusel2017NIPS} on ablating the choice of hierarchy on the Carla dataset. The first column is for training both LAE and LDM only with the coarse latent. The last column is our full model.
}
\vspace{-2mm}
\label{tab:ablation}
\end{table}

%% file: sections/conclusion.tex
\section{Conclusion}
\label{sec:conclusion}

We introduced \ourmodel\ (\ourmodelsrt), a generative model for complex 3D environments. 
\ourmodelsrt\ first constructs an expressive latent distribution by encoding input images into a 3D neural field representation which is further compressed into more abstract latent spaces. 
Then, our proposed hierarchical LDM is fit onto the latent spaces, achieving state-of-the-art performance on 3D scene generation.
\ourmodelsrt\ enables a diverse set of applications, including controllable scene generation and scene editing.
Future directions include exploring more efficient sparse voxel representations, training on larger-scale real-world data and learning to continously expand generated scenes.


%% file: sections/supple_embed.tex
\appendix
\addcontentsline{toc}{section}{Appendices}
\renewcommand{\thesection}{\Alph{section}}

\begingroup
\let\clearpage\relax
\onecolumn
\endgroup
\maketitle
\section*{\Large\selectfont{Supplementary Materials for \\ NeuralField-LDM: Scene Generation with Hierarchical Latent Diffusion Models}}

\thispagestyle{empty}

\title{}

\begingroup
\let\clearpage\relax
\onecolumn
\endgroup
\maketitle

\thispagestyle{empty}

\begin{figure*}[!thb]
  \centering
\includegraphics[width=1.0\textwidth]{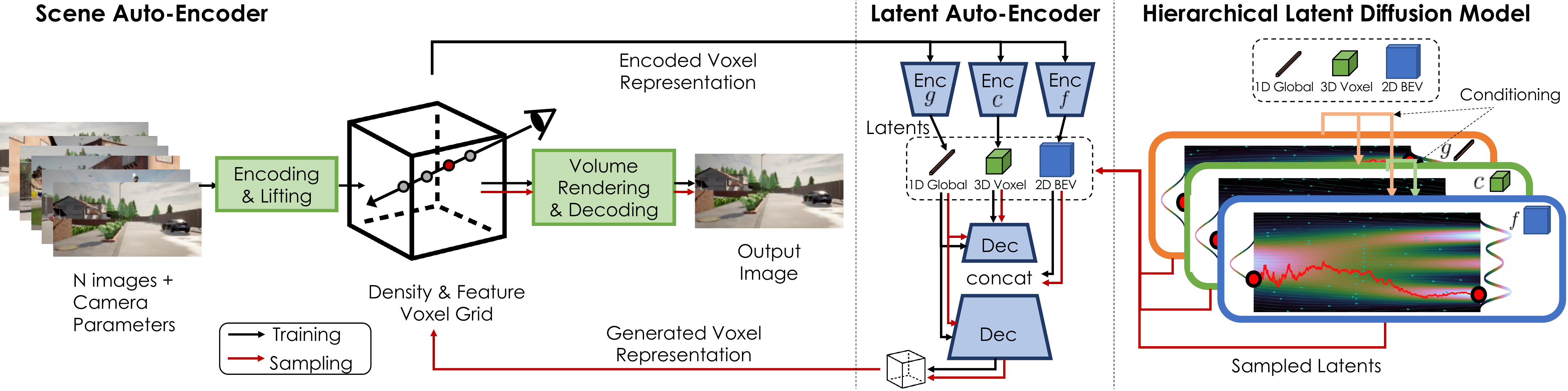}
   \caption{ 
   \textbf{Overview of \ourmodel}. 
   We put the model overview from the main text for reference. We first encode RGB images with camera poses into a neural field represented by density and feature voxel grids. We compress the neural field into smaller latent spaces and fit a hierarchical latent diffusion model on the latent space. Sampled latents can then be decoded into a neural field that can be rendered into a given viewpoint. 
    }
\label{fig:pipeline}
\end{figure*}

This supplementary document is organized as follows:
\begin{itemize}
  \item Section~\ref{sec:model_detail} includes additional model architecture and training details.
  \item Section~\ref{sec:results} includes additional qualitative results.
  \item We also include a supplementary video which has better visualizations for evaluating the 3D aspect of our model.
\end{itemize}

\section{Model Architecture and Training Details}
\label{sec:model_detail}
We use the convention $C\times Z\times X\times Y$ to denote a 3D tensor with $C$ channels, where the $Z$-axis points in the upward direction in 3D. Similarly,  $C\times H\times W$ denotes a 2D tensor with $C$ channels, with height $H$ and width $W$. In practice, we found that perceptual loss~\cite{zhang2018unreasonable} works better than L1 or L2 loss as an image reconstruction loss, and we use it throughout the paper as the loss function for image reconstruction.

\subsection{Scene Auto-Encoder}
\label{sec:scene_ae}

At every iteration, the scene auto-encoder takes as input $\{(i, \kappa, \rho)\}_{1..(N+M)}$ consisting of $N+M$ RGB images $i$ from the same scene along with their known camera posses $\kappa$ and depth measurements $\rho$, which can either be sparse (\eg Lidar points) or dense.
We use all cameras as supervision, but only use the first $N$ cameras as input to the model.

\textbf{Encoder Architecture}. We encode each of these $N$ images independently through an EfficientNet-B1~\cite{Tan2019ICML} encoder, replacing Batchnorm layers~\cite{Ioffe2015ICML} with Instance normalization~\cite{Ulyanov2016ARXIV}.
Additionally, the second and eigth blocks' padding and stride are modified to preserve spatial resolutions so that each image is downsampled by a factor of eight.
The EfficientNet-B1 head is replaced with a bilinear upsampling layer followed by a concatenation with the features before the last downsampling layers along the channel dimension and then two consecutive {Conv2d, ReLU, BatchNorm2d} layers, where the first Conv2D layer increases the channel dimension from $432$ to $512$.
These features are then processed by two seperate two-layer Conv networks that reduce the channel dimensions to $C$ and $D$, producing the feature and density values respectively for each pixel.
These feature and density values are used to define the frustum as explained in the Section 3.1 of the main text. We clamp the density to lie in $[-10,10]$ and apply the softplus activation function.
Each voxel in the density and feature voxel grids $V_{\texttt{Density}}$ and $V_{\texttt{Feat}}$ represents a region in the world coordinate system. 
For all datasets considered in this paper, we define the dimension of voxels to be $32\times128\times128\ (Z\times X\times Y)$. 
For VizDoom, each voxel represents a region of (4 game unit)$^3$.
For Replica, each voxel represents a region of $(0.125m)^3$.
For Carla, each voxel represents a region of $(0.75m)^3$.
For AVD, we use non-uniform voxel sizes. The voxels at the center have $0.2m$ side length, and the furthest voxels from the center have $1.6m$ horizontal side length and $2.4m$ vertical side length.

\textbf{Decoder Architecture}. We perform volumetric rendering, using the Mip-NeRF~\cite{barron2021mip} implementation on $V_{\texttt{Density}}$ and $V_{\texttt{Feat}}$  using $\{(\kappa)\}_{1..(N+M)}$ to get target features. These features are then fed through a decoder, using the blocks in StyleGAN2~\cite{Karras2020CVPRa} to produce output image predictions $\{\hat{i}\}_{1..(N+M)}$.
The decoder consists of ten StyledConv blocks, where the convolution operation of the fourth layer is replaced with a transposed convolution to upsample the features by a factor of 2.
A StyledConv block contains a style modulation layer~\cite{Karras2020CVPRa}, but we effectively skip the modulation process by feeding in a constant vector of $1$s.

\textbf{Training}. The parameters of the encoder and decoder are trained with an image construction loss, $||i-\hat{i}||$ across all $N+M$ inputs  with a coefficient of $1$. We also supervise the expected depth obtained from volumetric rendering with an MSE loss on pixels that contain a ground truth depth measurement weighted with a coefficient of $5$. Finally, we also add a regularization term on the sum over the entropy of all sampled opacity values from volumetric rendering to encourage very high or low values in the density voxels, weighted with a coefficient of $0.01$. 
For all models, we use the Adam optimizer with a learning rate of $0.0002$ and betas of $(0.,0.99)$. 
After training, we are able to further improve image quality by adding adversarial loss.
We use StyleGAN2's~\cite{Karras2020CVPRa} discriminator along with an R1 gradient regularization~\cite{Mescheder2018ICML}.
Furthermore, to capture the missing details from the encoding step while ensuring the distribution of training voxels does not diverge, we optionally perform a small number of additional per-scene optimization steps on the encoded voxels $V$. Specifically, for VizDoom, Replica and AVD, we perform 60 optimization steps by randomly sampling input views and reducing the image reconstruction loss, per encoded scene voxel.

\textbf{Camera Settings}. For VizDoom and Replica, we directly use the camera settings used in GSN~\cite{DeVries2021ICCV}. For Replica, we use all 100 consecutive frames per training sequence, and for VizDoom, as the area each sequence covers was too large for our voxel size, we chunk each training sequence into 50 consequtive frames. 
For Carla, at every iteration we sample a scene and a camera.
We sample $N+M=9$ consecutive frames from that scene and camera as our scene-encoder input, and randomly sample $N=6$ of those frames to input into the encoder.
We do this so we can obtain information across multiple timesteps, without incurring the memory cost of using all camera views at every iteration.
At inference time, for a given scene, we encode frames for all viewpoints at every timestep.
For AVD, we create a set of $5$ groups, each comprised of overlapping cameras.
We sample $N+M=8$ consecutive frames from a sampled scene and camera group as our scene-encoder input. We use $N=5$ fish-eye cameras as input to the encoder as they have the largest field-of-view and so the encoder does not have to learn to process different types of cameras. We use all cameras for the losses. We use histogram equalization on the input images. At inference time for AVD, we encode only the fish-eye cameras.

\subsection{Latent Voxel Auto-Encoder}
\label{sec:lae}
We concatenate $V_{\texttt{Density}}$ and $V_{\texttt{Feat}}$ along the channel dimension and use separate CNN encoders to encode the voxel grid $V$ into a hierarchy of three latents: 1D global latent $g$, 3D coarse latent $c$, and 2D fine latent $f$, as shown in Fig.~\ref{fig:pipeline}.
The intuition for this design is that $g$ is responsible for representing the global properties of the scene, such as the time of the day, $c$ represents coarse 3D scene structure, and $f$ is a 2D tensor with the same horizontal size $X\times Y$ as $V$, which gives further details for each location $(x,y)$ in bird's eye view perspective.  
We empirically found that 2D CNNs perform similarly to 3D CNNs while being more efficient, thus we use 2D CNNs throughout.
To use 2D CNNs for the 3D input $V$, we concatenate $V$'s vertical axis along the channel dimension and feed it to the encoders. 

\textbf{Encoder Architecture}. We use the building blocks of the encoder architecture from VQGAN\cite{Esser2021CVPR}.
Tables 1-3 contain the descriptions of the encoder architectures. 
Resblocks~\cite{He2016CVPR} contain two convolution layers and each conv layer has a group normalization~\cite{wu2018group} and a SiLU activation~\cite{elfwing2018sigmoid} prior to it. 
AttnBlocks are implemented as self-attention modules~\cite{Vaswani2017NIPS} and MidBlocks represent a block of \{ResBlock, AttnBlock, ResBlock\}. 
We add latent regularizations to avoid high variance latent spaces~\cite{Rombach2022CVPR}. 
For the 1D vector $g$, we use a small KL-penalty via the reparameterization trick~\cite{Kingma2014ICLR}, and for $c$ and $f$, we impose a vector-quantization~\cite{van2017neural,Esser2021CVPR} layer.
$c$ is quantized with a codebook containing 1024 entries, and $f$ is quantized with a codebook containing 128 entries.
Blocks that end with ``-CGN'' have group normalization layers replaced with conditional group normalization and they take in the global latent $g$ as the conditioning input.
Blocks that start with ``Unet-'' have a unet connection~\cite{ronneberger2015u} from their counterpart downsampling blocks that have the same feature dimension. For example, in the encoder for $f$, the Unet-ResBlocks take in the features of the first few ResBlocks and concatenate them to their input.

\begin{table}[h!]
  \begin{center}
    \begin{minipage}{.5\linewidth}
        \centering
        \begin{tabular}{c | c}
          \textbf{Layer}  & \textbf{Output dimension} \\
          \hline
          Input $V$ (3D) & $32\times32\times128\times128$ \\
          Concat $Z$-axis & $(32\times32)\times128\times128$ \\
          
          Conv2D 3$\times$3 & $128\times128\times128$\\
          6 $\times$ \{ResBlock  & \\
          ResBlock  & $128\times2\times2$ \\
          Conv2D 3$\times$3 stride 2\} &  \\
          ResBlock  & $128\times2\times2$ \\
          ResBlock  & $128\times2\times2$ \\
          AttnBlock  & $128\times2\times2$ \\
          MidBlock & $128\times2\times2$  \\
          Conv2D 2$\times$2 & $256\times1\times1$ \\
          Reparameterization (1D) & $128$
          \\

        \end{tabular}
        \caption{Encoder for the global latent $g$}
    \end{minipage}%
    \begin{minipage}{.5\linewidth}
        \centering
        \begin{tabular}{c | c}
          \textbf{Layer}  & \textbf{Output dimension} \\
          \hline
          Input $V$ (3D) & $32\times32\times128\times128$ \\
          Concat $Z$-axis & $(32\times32)\times128\times128$ \\
          
          Conv2D 3$\times$3 & $512\times128\times128$\\
          2 $\times$ \{ResBlock  & \\
          ResBlock  & $512\times32\times32$  \\
          Conv2D 3$\times$3 stride 2\} &  \\
          ResBlock  & $512\times32\times32$ \\
          ResBlock  & $512\times32\times32$ \\
          AttnBlock  & $512\times32\times32$ \\
          MidBlock & $512\times32\times32$  \\
          Conv2D 3$\times$3 & $32\times32\times32$ \\
          Split Z-axis & $4\times 8\times32\times32$ \\
          Quantization (3D) & $4\times 8\times32\times32$
          \\

        \end{tabular}
        \caption{Encoder for the coarse latent $c$}
    \end{minipage}%
  \end{center}
\end{table}

\begin{table}[h!]
  \begin{center}
    \begin{minipage}{.5\linewidth}
    
        \centering
        \begin{tabular}{c | c}
          \textbf{Layer}  & \textbf{Output dimension} \\
          \hline
          Input $V$ (3D) & $32\times32\times128\times128$ \\
          Concat $Z$-axis & $(32\times32)\times128\times128$ \\
          
          Conv2D 3$\times$3 & $256\times128\times128$\\
          2 $\times$ \{ResBlock  & \\
          ResBlock  & $256\times32\times32$ \\
          Conv2D 3$\times$3 stride 2\} &  \\
          MidBlock & $256\times32\times32$  \\
          Conv2D 3$\times$3 & $32\times32\times32$ \\
          Unet-MidBlock & $256\times32\times32$  \\
          2 $\times$ \{Unet-ResBlock-CGN  & \\
          Unet-ResBlock-CGN  &  \\
          ResBlock-CGN  & $256\times128\times128$\\
          Upsample2$\times$\} &  \\
          Conv2D 3$\times$3 & $4\times128\times128$ \\
          Quantization (2D) & $4\times128\times128$ \\
          \\

        \end{tabular}
        \caption{Encoder for the fine latent $f$}
    \end{minipage}%
    \begin{minipage}{.5\linewidth}
        
        \centering
        \begin{tabular}{c | c}
          \textbf{Layer}  & \textbf{Output dimension} \\
          \hline
          Input $c$ (3D) & $4\times8\times32\times32$ \\
          Concat $Z$-axis & $(4\times8)\times32\times32$ \\
        
          Conv2D 3$\times$3 & $512\times32\times32$\\
          MidBlock-CGN & $512\times32\times32$ \\
          ResBlock-CGN  & $512\times32\times32$ \\
          ResBlock-CGN  & $512\times32\times32$ \\
          ResBlock-CGN  & $512\times32\times32$ \\

          2 $\times$ \{ResBlock-CGN  & \\
          ResBlock-CGN  &  \\
          ResBlock-CGN  & $512\times128\times128$ \\ 
          Upsample2$\times$\} &  \\
          Combine $f$ & $512\times128\times128$  \\
          Conv2D 3$\times$3 & $1024\times128\times128$ \\
          Split Z-axis (3D)& $32\times32\times128\times128$  \\

        \end{tabular}
        \caption{Decoder of the latent auto-encoder}
    \end{minipage}%
  \end{center}
\end{table}

\textbf{Decoder Architecture}. The latent decoder architecture is presented in Table 4. 
It is similarly a 2D CNN, and takes $c$, concatenated along the vertical axis, as the initial input.
It also uses conditional group normalization layers with $g$ as the conditioning variable.
The fine latent $f$ is combined with an intermediate tensor in the decoder. 
This process is represented as ``Combine $f$'' in the table.
Specifically, we expand the channel dimension of $f$ to 128 with a 3$\times$3 Conv2D layer, and concatenate with the output tensor of the previous block. 
Then, it goes through three ResBlock-CGN layers to output a $512\times128\times128$ tensor. 
Finally, the tensor goes through a Conv2D layer and then is reshaped to the reconstructed voxel $\hat{V}$.

\textbf{Training}. The LAE is trained with the voxel reconstruction loss $||V-\hat{V}||$ along with the image reconstruction loss $||i-\hat{i}||$ where $\hat{i} = r(\hat{V},\kappa)$.
Note that the image reconstruction loss only helps with learning the LAE, and the scene auto-encoder is kept fixed.
For the voxel reconstruction loss, we divide $V$ into two groups. One group contains empty voxels that does not encode any information, and the other group have voxels filled in from the scene-autoencoding step in Section~\ref{sec:scene_ae}. The reconstruction loss is equally weighted between the two groups (\ie, we take the mean of the losses for the two groups separately and add them up). 
We use different weightings for $V_{\texttt{Density}}$ and $V_{\texttt{Feat}}$. The reconstruction loss for $V_{\texttt{Density}}$  is weighted 2.5$\times$ higher to encourage the model to reconstruct the geometry of the scene well.
We use a small KL coeffcient 2e-05 for $g$ which is multiplied to the KL loss.
We use a coefficient of 1.0 for the vector-quantization losses~\cite{van2017neural,Esser2021CVPR} for $c$ and $f$.
The image reconstruction loss is multiplied by 10.
We train the LAE with the Adam optimizer~\cite{Kingma2015ICML} with a learning rate of 0.0002.

\subsection{Hierarchical Latent Diffusion Models}

\textbf{Background on Denoising Diffusion Models}
Denoising Diffusion Models~\cite{sohl2015deep,ho2020ddpm,song2020score} (DDMs) are trained with denoising score matching to model a given data distribution $q(x_0)$.
DDMs sample a diffused input $x_t = \alpha_t x + \sigma_t \epsilon, \; \epsilon \sim \mathcal{N}(\mathbf{0}, \mathbf{I})$ from a data point $x \sim q(x_0)$ where $\alpha_t$ and $\sigma_t$ define a time $t$-dependent noise schedule.
The schedule is pre-defined such that the logarithmic signal-to-noise ratio $ \log(\alpha_{t}^2/\sigma_t^2)$ decreases monotonically.
Now, a neural network model $\psi$ is trained to denoise the diffused input by reducing the following loss
\begin{align}
\mathbb{E}_{x \sim q(x_0), t \sim p_{t}, \epsilon \sim \mathcal{N}(\mathbf{0}, \mathbf{I})} \left[\Vert y - \psi(x_t; t) \Vert_2^2 \right],
\label{eq:diffusionobjective}
\end{align}
where the target $y$ is either the sampled noise $\epsilon$ or $v = \alpha_t \epsilon - \sigma_t x$.
We use the latter target $v$ following \cite{salimans2022progressive} which empirically demonstrates faster convergence. 
$p_t$ denotes the distribution over time $t$ and we use a uniform discrete time distribution $p_{t} \sim \mathcal{U}\{0,1000\}$, following \cite{ho2020ddpm} .
We use the \emph{variance-preserving} noise schedule~\cite{song2020score}, for which $\sigma_t^2 = 1 - \alpha_t^2$.

\begin{figure*}[!thb]
  \centering
\includegraphics[width=0.6\textwidth]{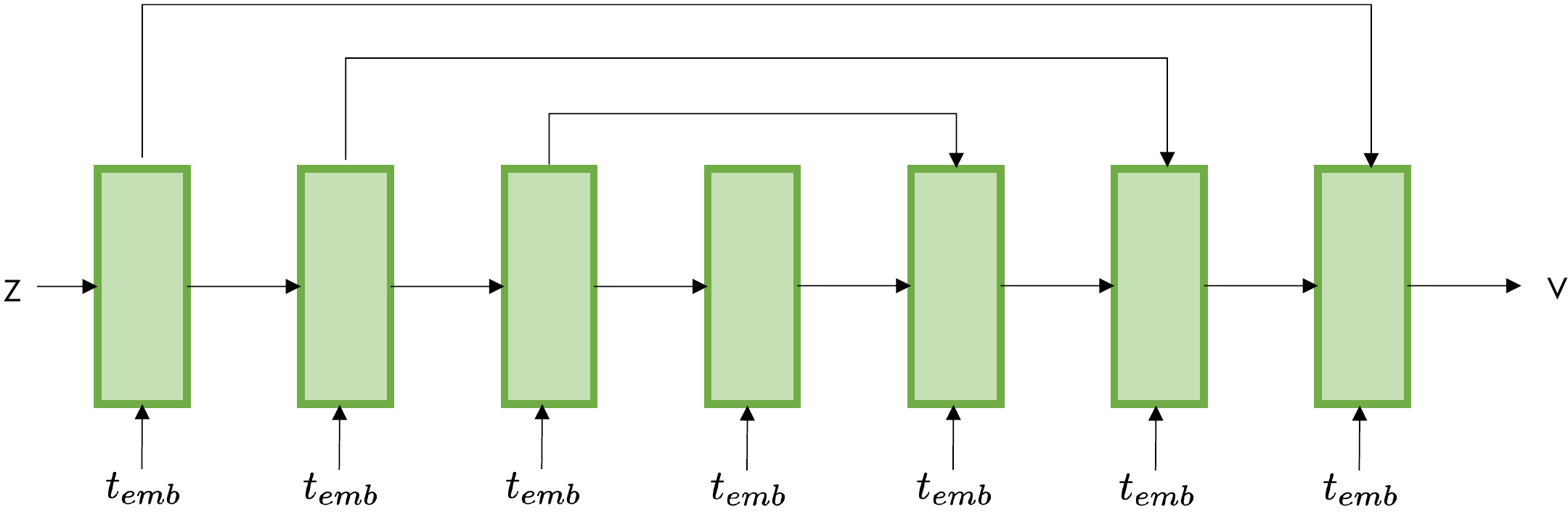}
   \caption{ 
   Architecture diagram of $\psi_g$. z is the input to the network and v is the output. The green blocks are the linear blocks with skip connections (Eq.~\ref{eq:lin_block}). The model is a 1D analogous version of the 2D Unet commonly used in 2D diffusion models.
    }
\label{fig:linearnet}
\end{figure*}

\textbf{Global Latent Diffusion Model}. The global LDM $\psi_g$ is implemented with linear blocks where each block is a residual block with skip connections:
\begin{equation}
\label{eq:lin_block}
\begin{split}
& h = linear(x) \\
& h_{emb} = linear(t_{emb}) \\
& h = h + h_{emb} \\
& h = linear(h) \\
& return \ linear(x) + h \\
\end{split}
\end{equation}
Here, $x$ is the input to the block and $t_{emb}$ is the timestep embedding for the diffusion time step $t$. We follow \cite{Rombach2022CVPR} to get the embedding.
We have $N$ such linear blocks. 
The inputs to the second half of the linear blocks are the concatenation of the previous block's output and the output of the corresponding first half of the linear block in a U-net fashion as depicted in Figure~\ref{fig:linearnet}.

As mentioned in the main text, the input to $\psi_g$ is both $g$ and the camera trajectory information which is flattened to 1D. 
For Carla and AVD, we implement $\psi_g$ as two separate networks that have the same architecture for the global latent and the camera trajectory. 
For Replica and VizDoom, we use a single network to model both the global latent and the camera trajectory as they are highly correlated (\eg we found that each global latent in Replica represents a scene in the training dataset and trajectories should be sampled within the given scene, as otherwise, it could go out of the bound of the scene).
Table~\ref{tab:global_ldm} contains the hyperparameter choices for $\psi_g$.

\begin{table}
\centering
\begin{tabular}{c|c|c|c|c}
    \toprule
     & VizDoom  & Replica & Carla & AVD \\
    \midrule
   Global Latent Dimension & 128 & 128 & 128 & 128 \\
   Trajectory Dimension & 200 & 400 & 18 & 24 \\
   Number of Linear Blocks & 10 &  10  & 10 &  6\\
    Channel dimension & 2048  & 2048 & 512 & 2048\\
    Learning Rate & 5e-05  & 5e-05 & 5e-05 & 5e-05\\
    
    \bottomrule
\end{tabular}
\caption{
Hyperparameters for $\psi_g$. Each training sequence in VizDoom consists of 50 timesteps, each with three-dimensional $(x,y,z)$ location information and one-dimensional yaw information totalling 200 dimensions per trajectory. Similarly, Replica has 100 timesteps, totalling 400 dimensions per trajectory. Carla has the same $z$ location for the Z-axis across different timesteps, so we only model the $(x,y)$ trajectory information from nine consecutive timesteps. For AVD, we model all three $(x,y,z)$ translation parameters across eight timesteps, totalling 24 dimensions per trajectory. 
}
\label{tab:global_ldm}
\end{table}

\textbf{Coarse and Fine Latent Diffusion Model}.
$\psi_c$ and $\psi_f$ adopt the U-net architecture~\cite{ronneberger2015u} and closely follow the 2D Unet architecture used in\cite{Rombach2022CVPR}. 
The input to $\psi_c$ is 3D but we concatenate it along the $Z$-axis and use the 2D Unet architecture without introducing 3D components. The output is split along the channel dimension to recover the 3D output shape.
$\psi_f$ also takes in $c$ as the conditioning input. We first concatenate $c$ along the $Z$-axis, making its shape $32\times 32\times 32$, and then interpolate it to match the spatial dimension of $f$ to be a tensor with shape $32\times 128\times 128$. Finally, the interpolated $c$ is concatenated to $f$ (so the shape of the concatenated tensor is $36\times 128\times 128$) and fed into the Unet model whose output matches the shape of $f$, $4\times 128\times 128$.
Similar to $\psi_g$, $\psi_c$ and $\psi_f$ also take the timestep embedding $t_{emb}$ for the sampled diffusion time step $t$.
Table~\ref{tab:coarse_ldm} and Table~\ref{tab:fine_ldm} contain the hyperparameter settings for $\psi_c$ and $\psi_f$, respectively.

The cross attention layers in \cite{Rombach2022CVPR} are equivalent to self-attention layers if no extra conditioning information is given. For Bird's eye view (BEV) segmentation conditioned models, we additionally train a 2D convolution encoder network that takes in the segmentation map with size $\mathbb{R}^{3\times 128\times 128}$ and produces a BEV embedding with size $\mathbb{R}^{256\times 32\times 32}$.
This BEV embedding is fed into the cross attention layers for conditional synthesis for $\psi_c$ and $\psi_f$. For $\psi_g$, we take the mean of the embedding across the spatial dimension, and concatenate with the timestep embedding that goes into the linear blocks.

\textbf{Training}.
We follow \cite{Rombach2022CVPR} for the choice of diffusion steps (1000), noise schedule (linear), and optimizer (AdamW~\cite{Loshchilov2019ICLR}) for all experiments.
For sampling, we use the DDIM sampler~\cite{song2020denoising} with 250 steps.

\begin{table}
\centering
\begin{tabular}{c|c|c|c|c}
    \toprule
     & VizDoom  & Replica & Carla & AVD \\
    \midrule
    Input Shape & $4\times8\times32\times32$ & $4\times8\times32\times32$ & $4\times8\times32\times32$ & $4\times8\times32\times32$ \\
    Channels & 224 &  128  & 288 &  256\\
    Channel Multiplier & 1,2,3,4 & 1,2,3,4 & 1,2,3,4 & 1,2,3,4 \\
    Attention Resolutions  & 4,8,16  & 4,8,16 & 4,8,16 & 4,8,16\\
    Learning Rate & 6.4e-05  & 6.4e-05 & 6.4e-05 & 6.4e-05\\
    \bottomrule
\end{tabular}
\caption{
Hyperparameters for $\psi_c$. Channels denote the base number of channels. Each group of layers (four groups in our case as indicated by the number of channel multipliers) in the Unet (see \cite{Rombach2022CVPR} for further details) have the number of channels equal to the base channels multiplied by the corresponding channel multiplier. Attention layers are applied at the specified 2D spatial resolutions. The tensor with the smallest spatial resolution in the Unet has $4\times 4$ spatial resolution.
}
\label{tab:coarse_ldm}
\end{table}

\begin{table}
\centering
\begin{tabular}{c|c|c|c|c}
    \toprule
     & VizDoom  & Replica & Carla & AVD \\
    \midrule
    Input Shape & $4\times128\times128$ & $4\times128\times128$ & $4\times128\times128$ & $4\times128\times128$ \\
    Channels & 128 &  128  & 288 &  512\\
    Channel Multiplier & 1,2,2,2,4,4 & 1,2,2,2,4,4 & 1,2,2,2,4,4 & 1,1,1,1,1,1 \\
    Attention Resolutions  & 16,32,64  & 16,32,64 & 16,32,64 & 8,16,32 \\
    Learning Rate & 6.4e-05  & 6.4e-05 & 6.4e-05 & 6.4e-05\\
    \bottomrule
\end{tabular}
\caption{
Hyperparameters for $\psi_f$. Channels denote the base number of channels. Each group of layers (six groups in our case as indicated by the number of channel multipliers) in the Unet (see \cite{Rombach2022CVPR} for further details) has the number of channels equal to the base channels multiplied by the corresponding channel multiplier. Attention layers are applied at the specified 2D spatial resolutions. The tensor with the smallest spatial resolution in the Unet has a $4\times 4$ spatial resolution.
}
\label{tab:fine_ldm}
\end{table}

\subsection{Post-Optimizing Generated Neural Fields}
\label{sec:post_opt}

Given a set of voxels $V$, obtained either through sampling or by encoding a set of views, we are able to increase the quality of $V$ through post-optimization using SDS loss as shown in Figures \ref{fig:av_sample1} and \ref{fig:av_sample2}. 
For the entire optimization, we use a fixed set of camera parameters $\{\kappa\}_{1\dots N}$ sampled from the training dataset scene as the base camera position where, for AVD, $N=6$ and all intrinsic matrices are replaced with the camera intrinsic parameters from the non-fisheye left-facing camera. 
At every iteration, we uniformly sample a translation offset in both the forwards and sideways directions between $-3$ and $3$ metres as well as a rotation offset about the $Z$-axis uniformly between $-10$ and $10$ degrees. 
We apply these offsets to $\{\kappa\}_{1\dots N}$ to obtain $\{\hat{\kappa}\}_{1\dots N}$, and render out images $\hat{i} = r(V, \hat{\kappa})$ for each viewpoint.
We then either use random cropping or left/right cropping to make the aspect ratio square and bilinearly resize $\hat{i}$ to $512 \times 512$ resolution, matching the required input dimensions for the diffusion model.
We obtain the gradient for the voxels using Equation 7 in the main text, leaving the decoder parameters fixed.

We use an off-the-shelf latent diffusion model~\cite{Rombach2022CVPR}, finetuned to condition on CLIP image embeddings~\cite{Radford2021ARXIV}.
We train with negative guidance, as detailed in Section~\ref{sec:negative_guidance}.
For the positive conditioning, we sample $23k$ images from the front, left and right facing non-fisheye cameras from our dataset and take the average of their CLIP image embeddings. 
For the negative conditioning, we sample $80$ voxels from our model and use the average CLIP image embeddings from $23k$ images rendered from the voxels using the same camera jitter distribution we use for post-optimization.
We attempted to use classifier-free guidance without the negative conditioning, but found the outputs to be blurry as seen in Figure~\ref{fig:voxel_ablation}.
At every update, we uniformly sample the noising timestep, $t \in [20, 200]$, independently for each image in the batch.

We train with a batch-size of $3$ and a gradient accumulation of $2$ steps, fixing the cameras in the even updates and odd updates so every gradient step contains updates from every camera view exactly once.
We use the Adam optimizer with a learning rate of 1e-3, betas of $(0.9,0.99)$ and epsilon set to 1e-15.
We optimize a single scene for $20k$ iterations, taking approximately $13$ hours on a single $V100$ GPU, but also see drastic quality improvements after $2k$ iterations.

We note that as seen in Figure \ref{fig:voxel_ablation}, having a voxel initialization sampled or encoded from our model is critical to the success of post-optimization.

\subsubsection{Negative-guidance}
\label{sec:negative_guidance}

Let $y,y'$ be positive conditioning (\eg dataset image) and negative conditioning (\eg sampled images with artifacts), respectively, and $x$ a diffusion-step sample. Intuitively, we want to sample the diffusion model so that $p(x | y)$ is high and $p(x | y')$ is low. Thus, we want to sample from $\frac{p(x|y)^{\alpha}}{p(x|y')}$ where $\alpha$ trades off the importance of sampling towards $y$ and away from $y'$. We see then that:

$$\nabla_x \log \frac{p(x|y)^{\alpha}}{p(x|y')} =  \alpha \nabla_x \log p(x|y)  - \nabla_x \log p(x|y')$$

which is equal to classifier free guidance with $\gamma = 2$, $\alpha=\gamma=2$ and the unconditional embedding replaced with $y'$.

For reference, classifier-free guidance is defined as: 

$$\gamma \nabla_x \log p(x|y)  + (1-\gamma) \nabla_x \log p(x)$$

Empirically, we implement classifier-free guidance and replace the uncondtional embedding with $y'$ which, as shown below, is equivalent to setting $\alpha= \frac{\gamma}{\gamma-1}$ and multiplying the gradient by $(\gamma-1)$:

\begin{align*}
\gamma \nabla_x \log p(x|y)  + (1-\gamma) \nabla_x \log p(x|y') &= \gamma \nabla_x \log p(x|y)  - (\gamma-1) \nabla_x \log p(x|y') ) \\
&= (\gamma-1)( \frac{\gamma}{\gamma-1} \nabla_x \log p(x|y)  - \nabla_x \log p(x|y') ) \\
&= (\gamma-1)\nabla_x \log \frac{p(x|y)^{\frac{\gamma}{\gamma-1}}}{p(x|y')}
\end{align*}

\newpage

\section{Additional Results}
\label{sec:results}

\subsection{More Ablations}
\begin{figure}[t]
\begin{center}
    \includegraphics[width=0.7\linewidth]{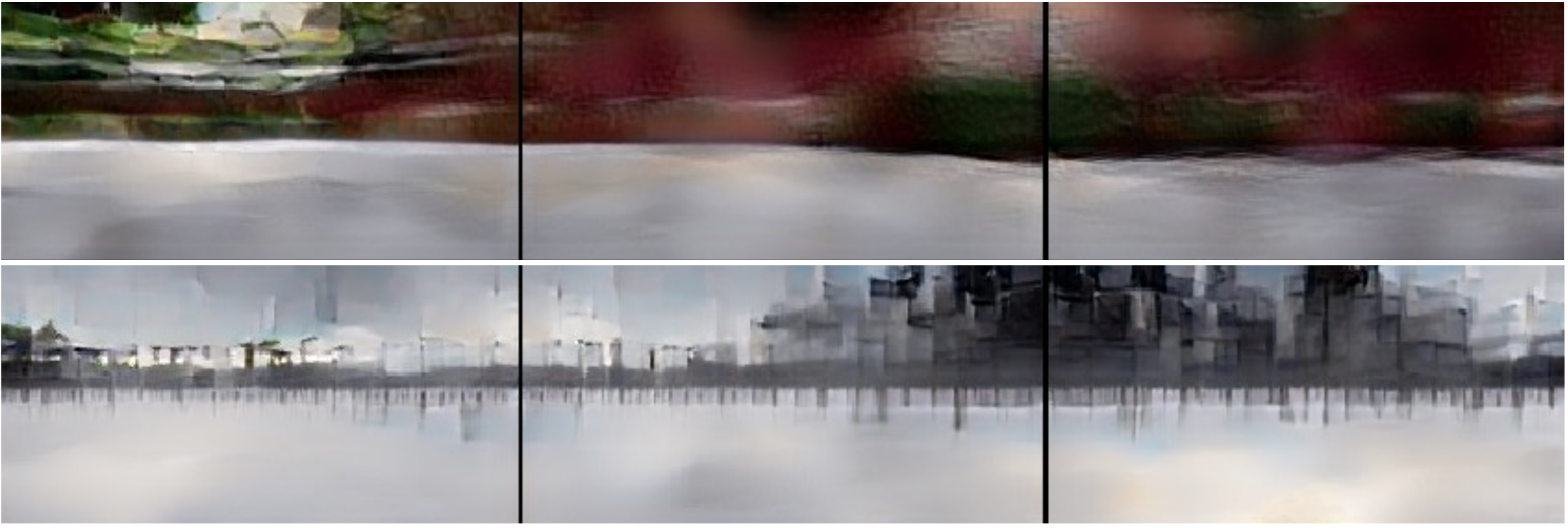}
\end{center}
   \caption{\small \textbf{Directly fitting a diffusion model without compression with latent auto-encoder is challenging.} Each row is a sample from a diffusion model trained directly on the $128\times128\times32$ grids from the first stage autoencoder.}
\label{fig:fullvoxel}
\end{figure}

\textit{\textbf{(1) Scene Encoder:}} the voxel size used by the scene encoder is crucial in capturing details of the scene. If we use larger voxel size and encoder frustum size, the voxel would be able to contain more pixel-level detail and consequently output better quality images.
However, this comes with the disadvantage that modelling such high-dimensional voxel space with a generative model becomes challenging. 
In Fig.~\ref{fig:fullvoxel}, we show samples from a diffusion model fit to our first-stage voxels for Carla. We hypothesize that current DMs cannot perform well on very high dimensional data, highlighting the importance of our hierarchical latent space.
Tab.~\ref{tab:ablation_ae} reports perceptual loss on reconstructed output viewpoints. We concluded that $128\times128\times32$ provides a satisfactory output quality while still being small enough for the consequent stages to model and to not consume excessive GPU memory.

\textit{\textbf{(2) Latent Encoder:}} as mentioned, having larger voxels gives better reconstruction, but fitting a generative model becomes more challenging. Therefore, our latent auto-encoder compresses voxels into smaller latents, and in Tab.~\ref{tab:ablation_lae}, we report how downsampling factors (for the coarse 3D latent) in the encoder affect the voxel reconstruction quality. We found that $ds=4$ gives a good compromise between having a low reconstruction loss and a latent size small enough to fit a diffusion model.

\begin{figure}[t]
\begin{center}
    \includegraphics[width=0.7\linewidth]{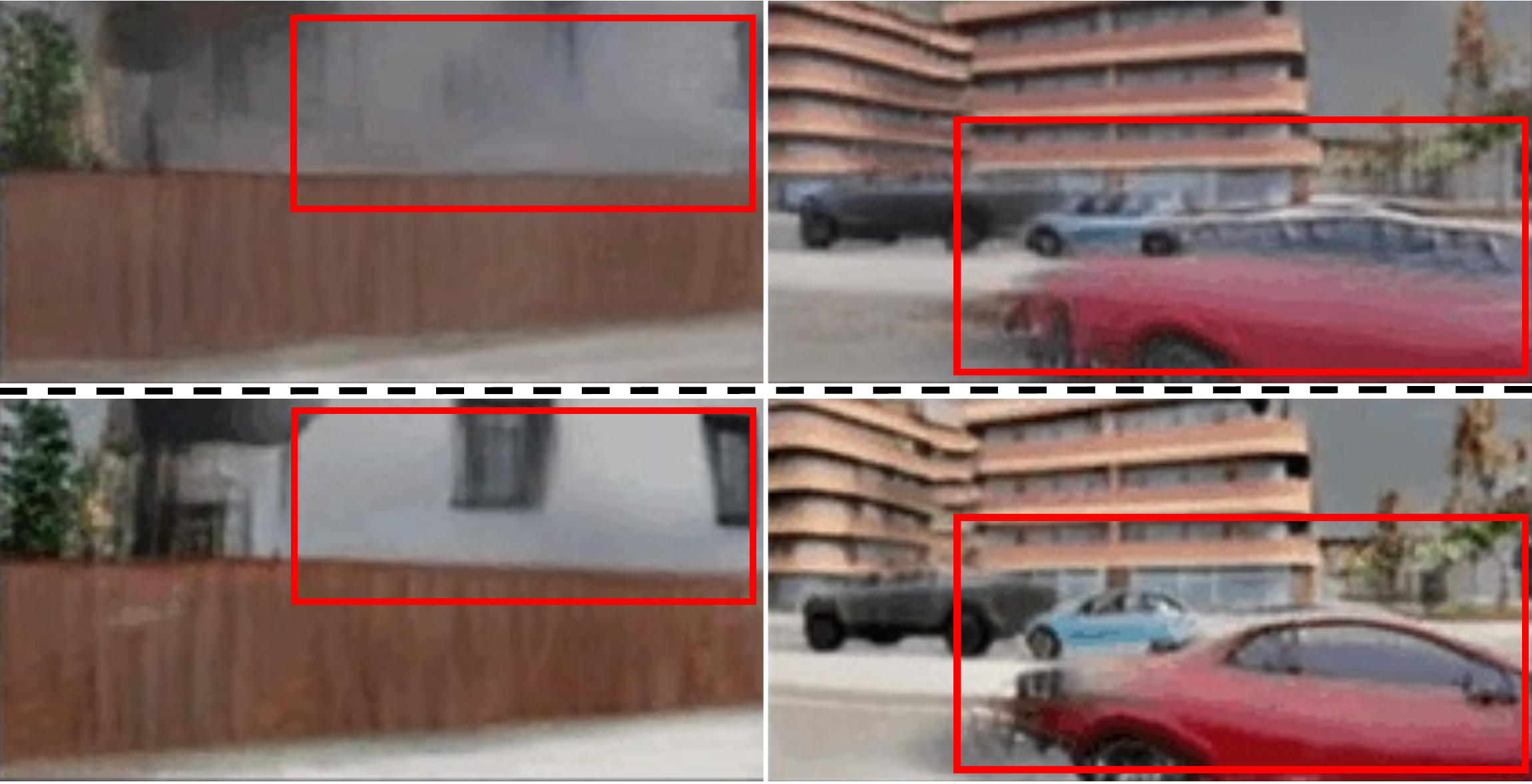}
\end{center}
   \caption{\small Renderings from the scene autoencoder. \emph{Top row}: without explicit density \& feature grids, \emph{Bottom row}: the full model.}
\label{fig:desntiyvoxel}
\end{figure}

\textit{\textbf{(3) Explicit Density:}} in Fig. \ref{fig:desntiyvoxel}, we show that having explicit feature and density grids outperforms implicitly inferring density from the voxel features with an MLP.
Our encoder explicitly predicts the occupancy of each frustum entry before merging frustums across multiple views and thus prevents incorrect feature mixing due to occlusions that can happen if frustums are merged with naive mean-pooling without accounting for occupancy.
Implicit depth prediction similar to Lift-Splat~\cite{Philion2020ECCV} can also account for occlusion but this requires an additional density prediction step for volume rendering which we avoid by predicting densities directly from each view.

\textit{\textbf{(4) Sampling Steps:}} sampling with a larger number of steps only marginally improved FID - 50/37.18, 125/36.74, 250/35.69 (\# steps/FID with DDIM sampler $\eta=1.0$).

\begin{table}
\centering
\begin{tabular}{c|ccc}
    \toprule
     & $32\times32\times8$  &  $64\times64\times16$ & $128\times128\times32$  \\
    \midrule
    Percept. Loss ($\downarrow$) & 0.3508 & 0.2688 & \textbf{0.2237}  \\
    \bottomrule
\end{tabular}
\caption{
Ablation of the voxel dimensions of the scene autoencoder.
We report the validation perceptual loss.
}
\label{tab:ablation_ae}
\end{table}

\begin{table}
\centering
\begin{tabular}{c|ccc}
    \toprule
     & $ds=16$  &  $ds=8$ & $ds=4$  \\
    \midrule
    Vox. Recon Loss ($\downarrow$) & 0.6076 & 0.5949 & \textbf{0.4915}  \\
    \bottomrule
\end{tabular}
\caption{
Ablation of the downsampling factors ($ds$) of the latent autoencoder. We report the validation
voxel reconstruction loss.
}
\label{tab:ablation_lae}
\end{table}

\subsection{Generated Scenes}

\begin{figure*}[!thb]
  \centering
\includegraphics[width=0.77\textwidth]{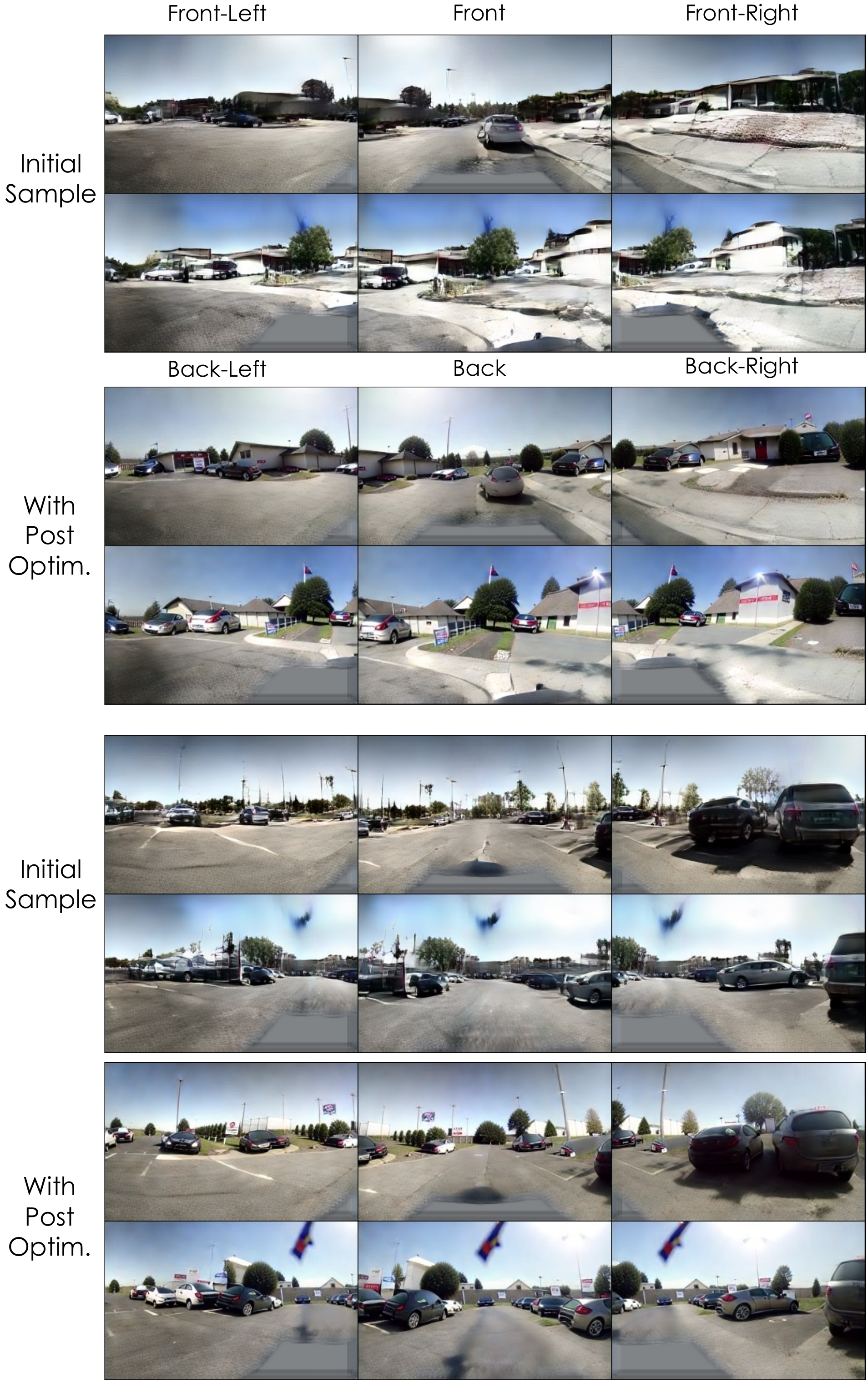}
   \caption{Additional generated samples on AVD. Each initial sample is further improved with post-optimization  (Section~\ref{sec:post_opt}).
    }
\label{fig:av_sample1}
\end{figure*}

\begin{figure*}[!thb]
  \centering
\includegraphics[width=0.77\textwidth]{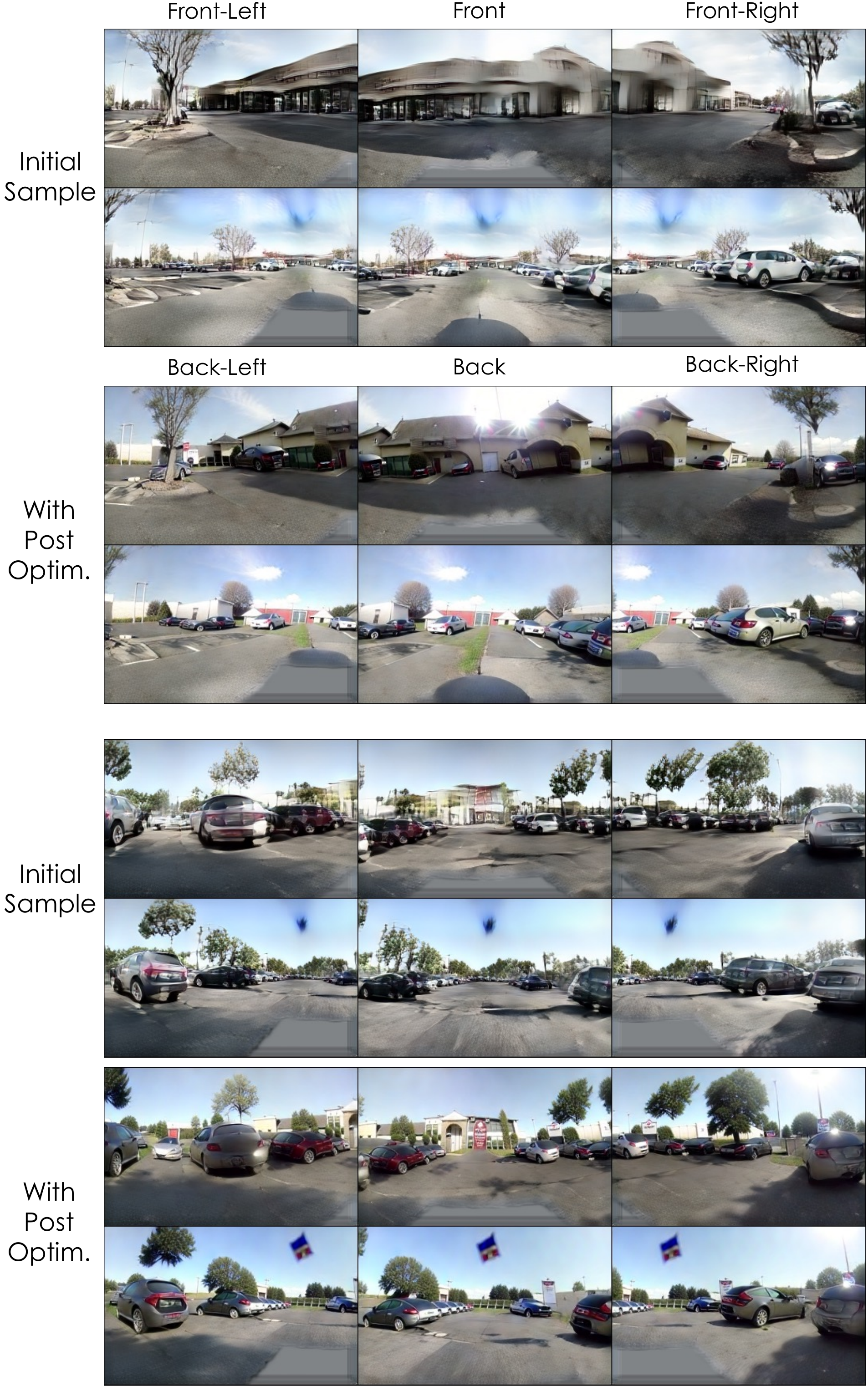}
   \caption{Additional generated samples on AVD. Each initial sample is further improved with post-optimization (Section~\ref{sec:post_opt}). 
    }
\label{fig:av_sample2}
\end{figure*}

\begin{figure*}[!thb]
  \centering
\includegraphics[width=0.7\textwidth]{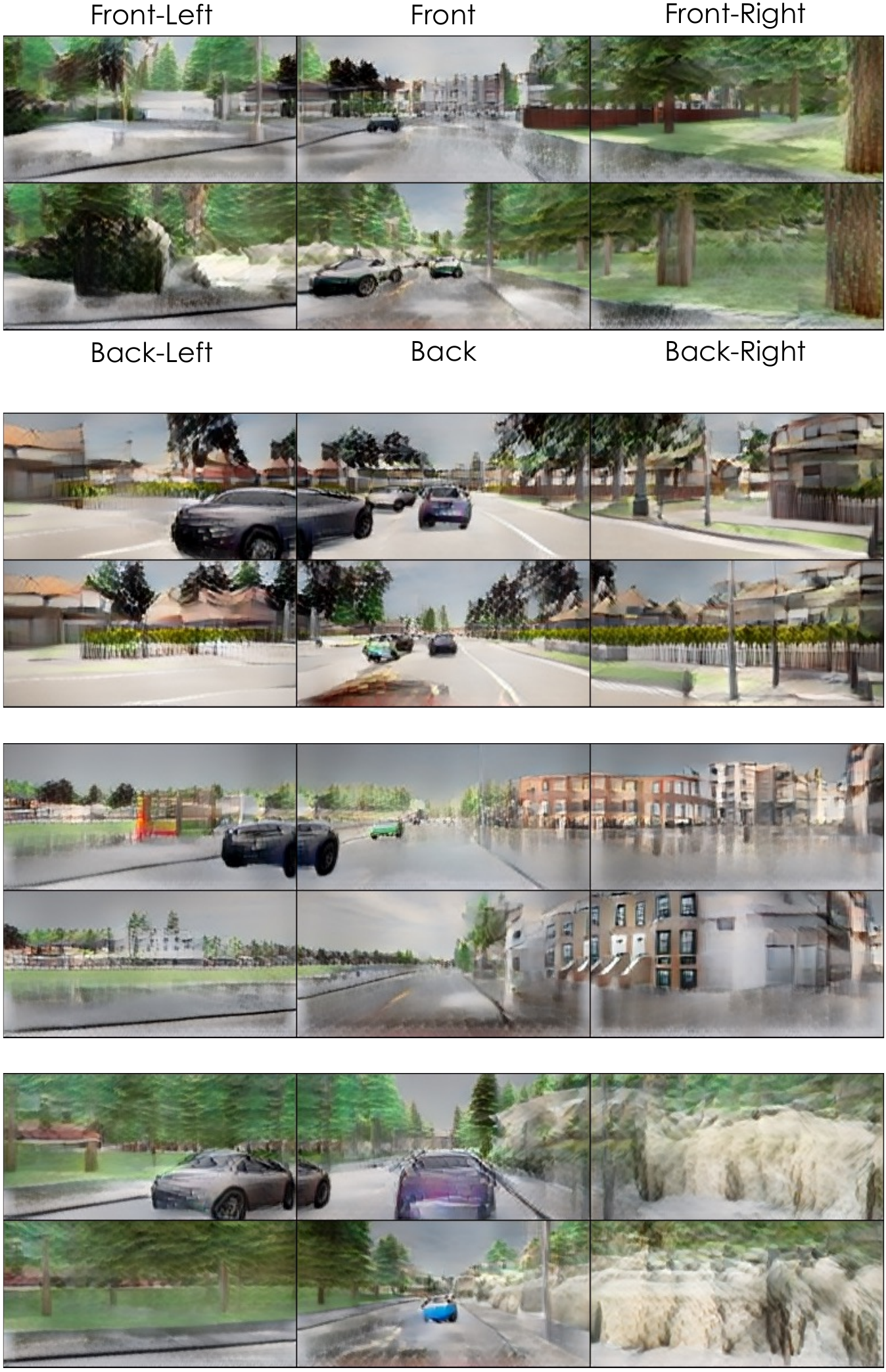}
   \caption{Additional generated samples on Carla.
    }
\label{fig:supp_carla_sample1}
\end{figure*}

\begin{figure*}[!thb]
  \centering
\includegraphics[width=0.7\textwidth]{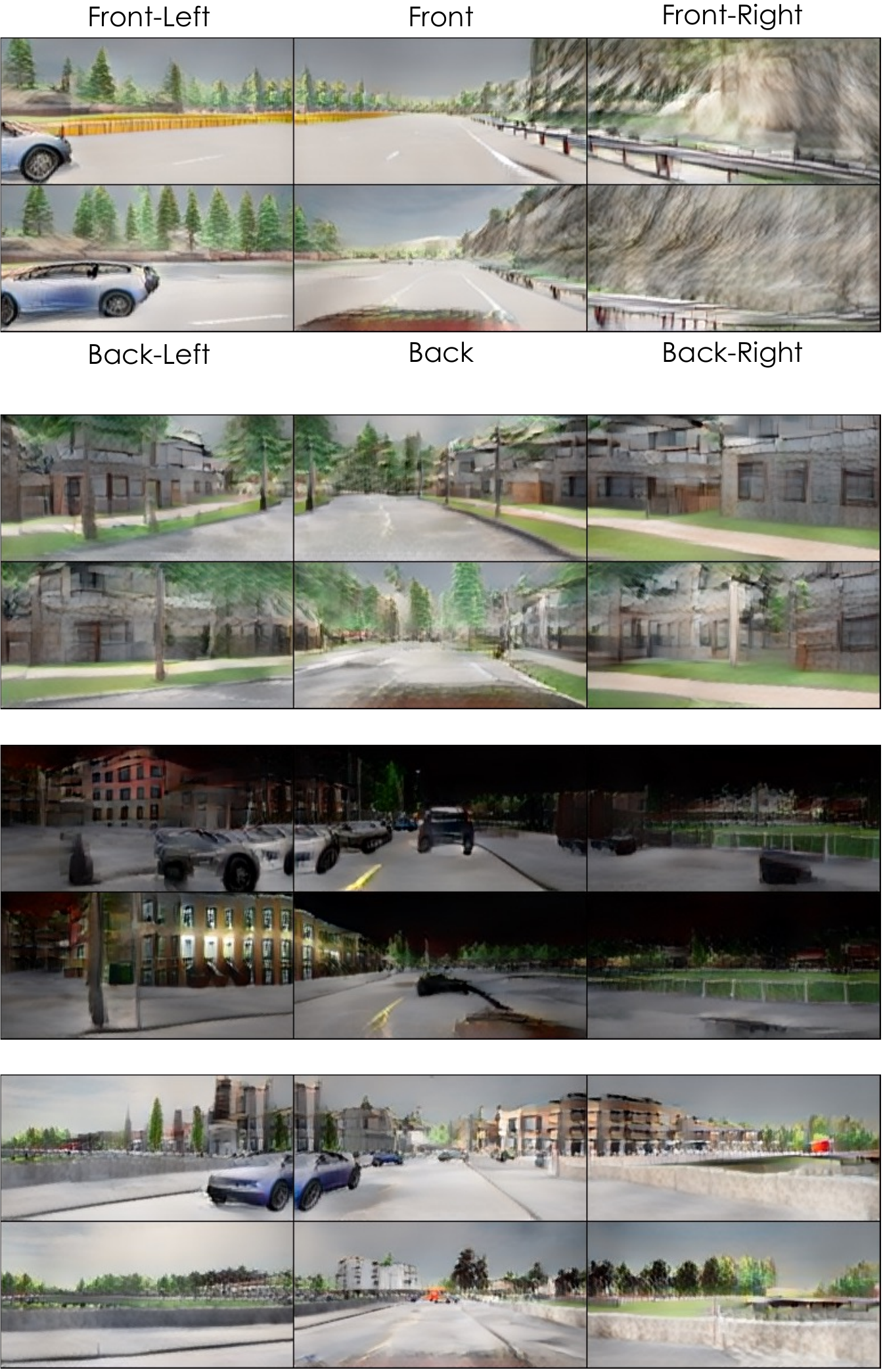}
   \caption{Additional generated samples on Carla.
    }
\label{fig:supp_carla_sample2}
\end{figure*}

\begin{figure*}[!thb]
  \centering
  \hspace{-5mm}
\includegraphics[width=0.95\textwidth]{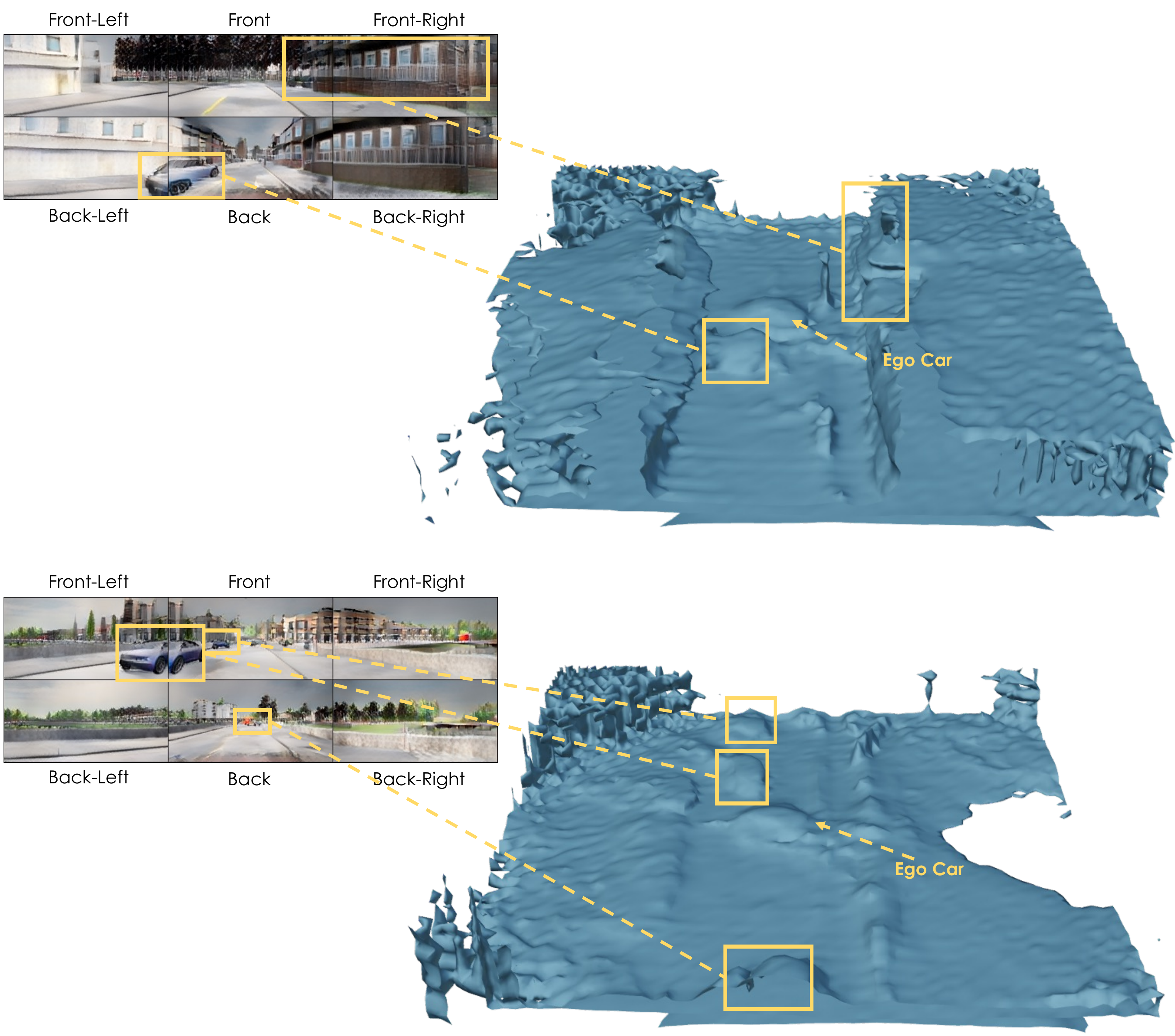}
   \caption{We run marching-cubes~\cite{Lorensen1987SIGGRAPH} on the density voxels to visualize the geometry of the samples generated by \ourmodelsrt.
    }
\label{fig:supp_mesh}
\end{figure*}

We provide additional generated samples on AVD in Figures~\ref{fig:av_sample1} and \ref{fig:av_sample2}.
For Carla, we provide samples in Figure~\ref{fig:supp_carla_sample1} and \ref{fig:supp_carla_sample2}. Figure~\ref{fig:supp_mesh} contains visualizations of 3D meshes obtained by running marching-cubes~\cite{Lorensen1987SIGGRAPH} on samples.

\clearpage

\clearpage


\clearpage

\subsection{Stylization using Score Distillation Sampling (SDS) loss}
\label{sec:style_SDS}
In addition to using SDS loss to post-optimize our voxels for quality, we can also use it to modify the style of a  given scene.
Given a desired target style (\eg a medieval castle), we first generate a dataset of target (positive) and source (negative) images using one of two methods:
\begin{itemize}
    \item \textbf{Image translation}: We use stable diffusion~\cite{Rombach2022CVPR} for text-guided image-to-image translation as introduced by SDEdit~\cite{meng2021sdedit}. Specifically, we autoencode scenes from our dataset to contruct a set of reconstructed images which we use as the source images. 
    We then run the image to image translation on the source's matching dataset images, using a strength of $0.4$ and guidance scale of $10$, using the text of the target style to get target images.
    We repeat this for $500$ images and take the average of the source images' CLIP embeddings and target images' CLIP embeddings as $y'$ and $y$ used in negative guidance respectively.
    \item \textbf{Scraping}: We use the same negative conditioning $y'$ as we do for quality post-optimization. For, $y$, we search and download $100$ images from the internet with the target query, manually filter these images for relevance and take the average CLIP embedding.
\end{itemize}

\begin{figure*}[!thb]
  \centering
  \hspace{-15mm}
\includegraphics[width=0.83\textwidth]{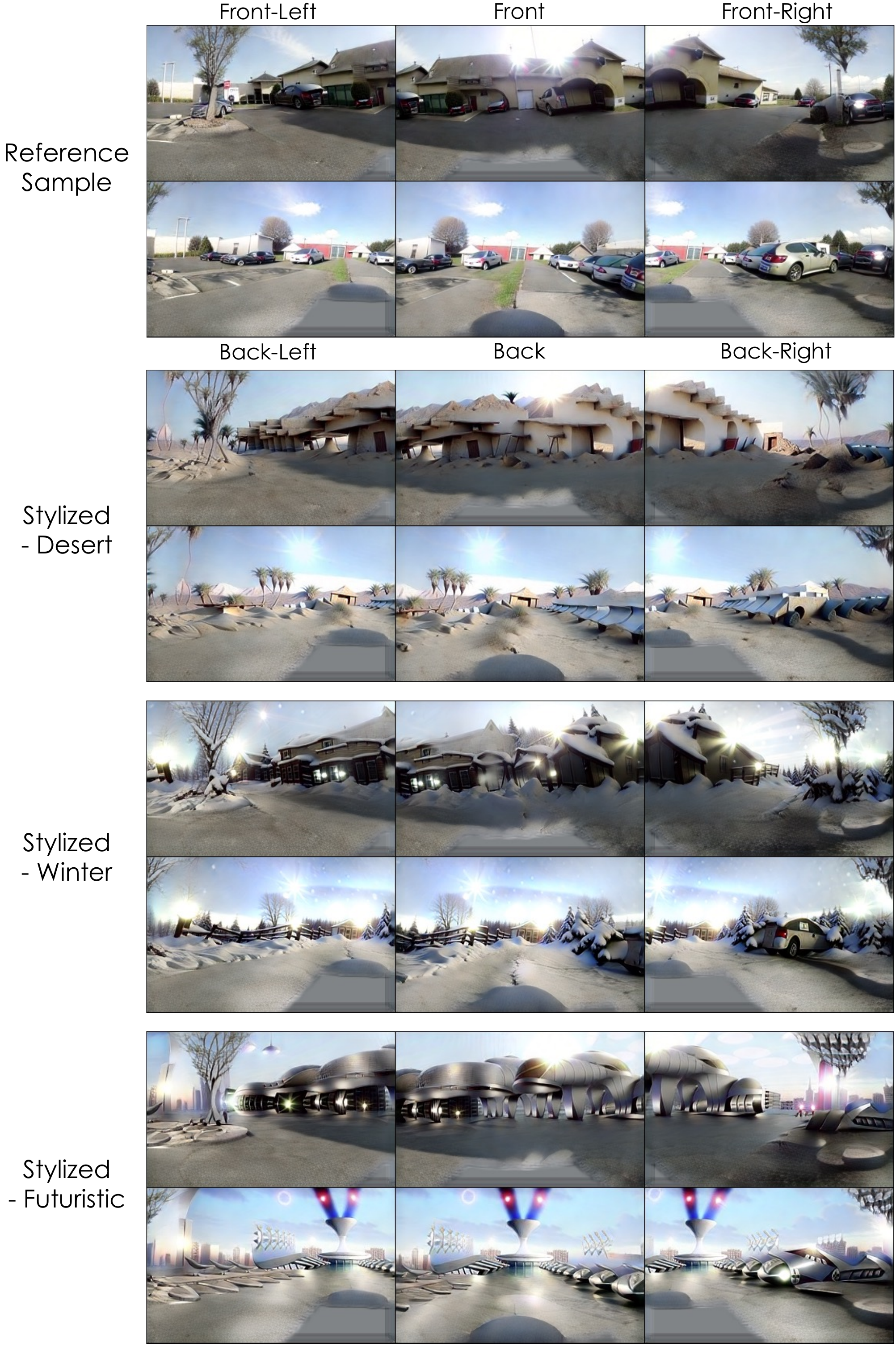}
   \caption{Additional stylized samples.  All stylized samples start the post-optimization step from the same initial sample.
    }
\label{fig:supp_style1}
\end{figure*}

\begin{figure*}[!thb]
  \centering
  \hspace{-15mm}
\includegraphics[width=0.83\textwidth]{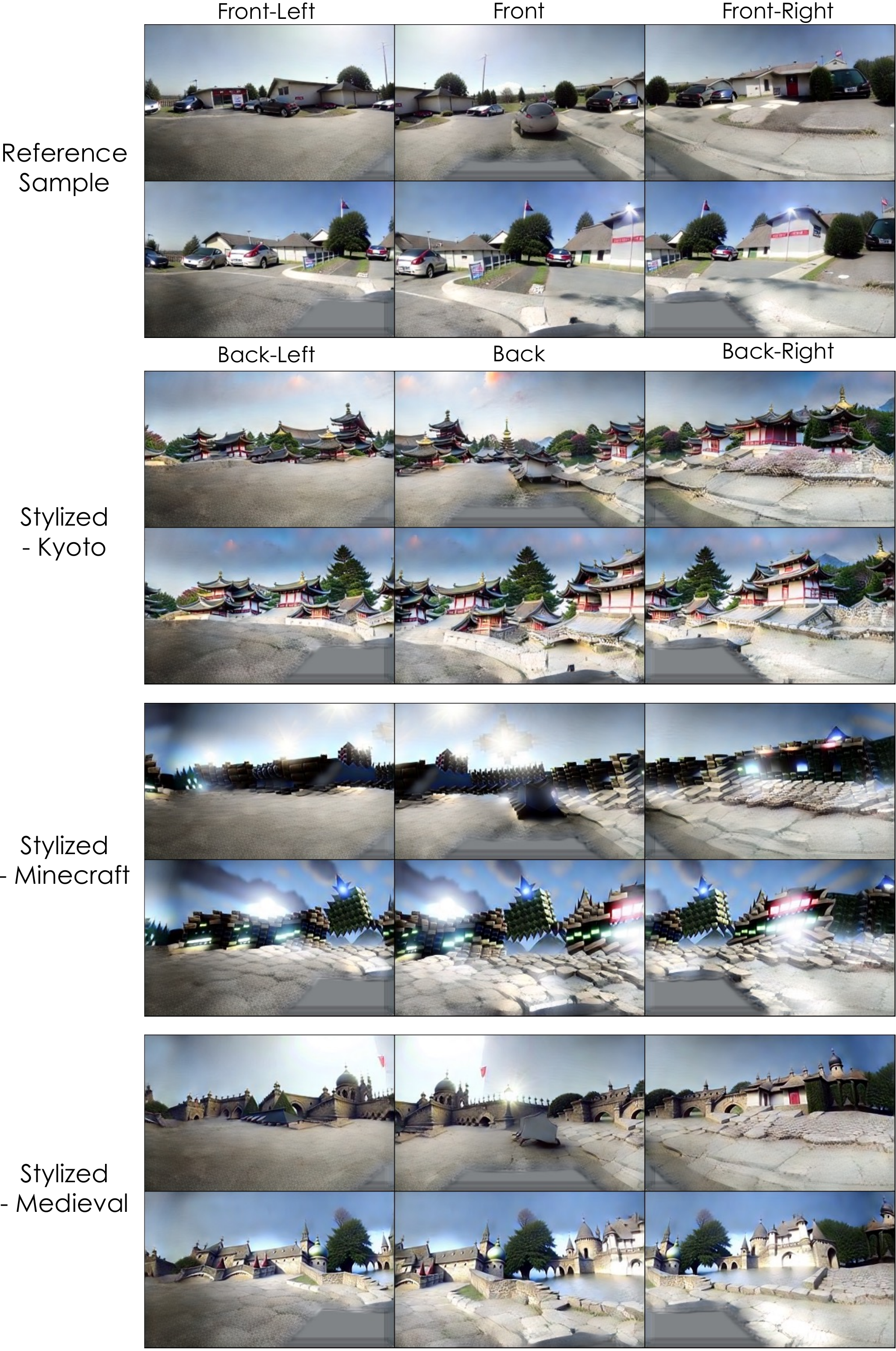}
   \caption{Additional stylized samples. All stylized samples start the post-optimization step from the same initial sample.
    }
\label{fig:supp_style2}
\end{figure*}

\begin{figure*}[!thb]
  \centering
  \hspace{-15mm}
\includegraphics[width=0.83\textwidth]{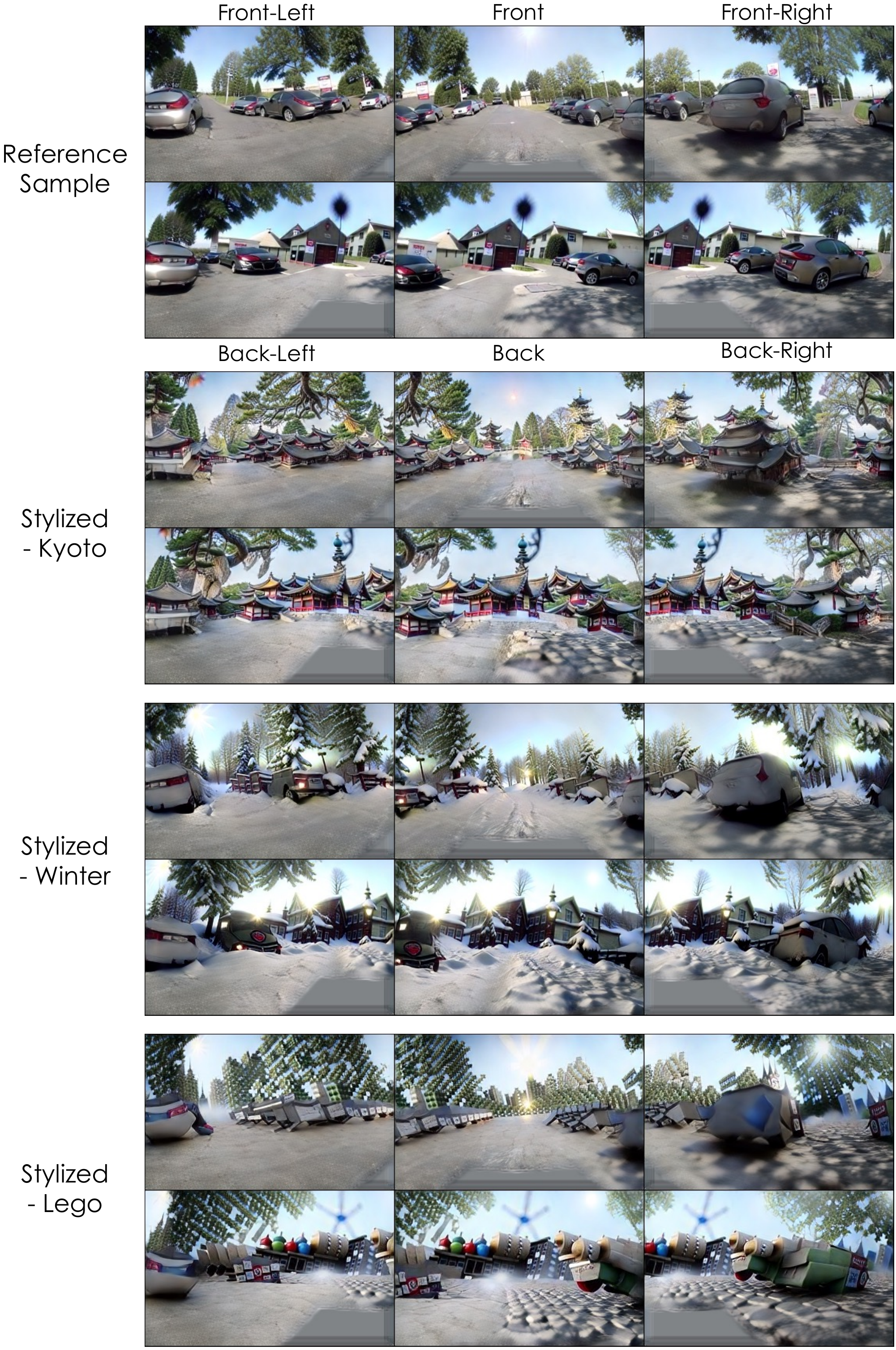}
   \caption{Additional stylized samples. All stylized samples start the post-optimization step from the same initial sample.
    }
\label{fig:supp_style3}
\end{figure*}

\begin{figure*}[!thb]
  \centering
  \hspace{-5mm}
\includegraphics[width=0.83\textwidth]{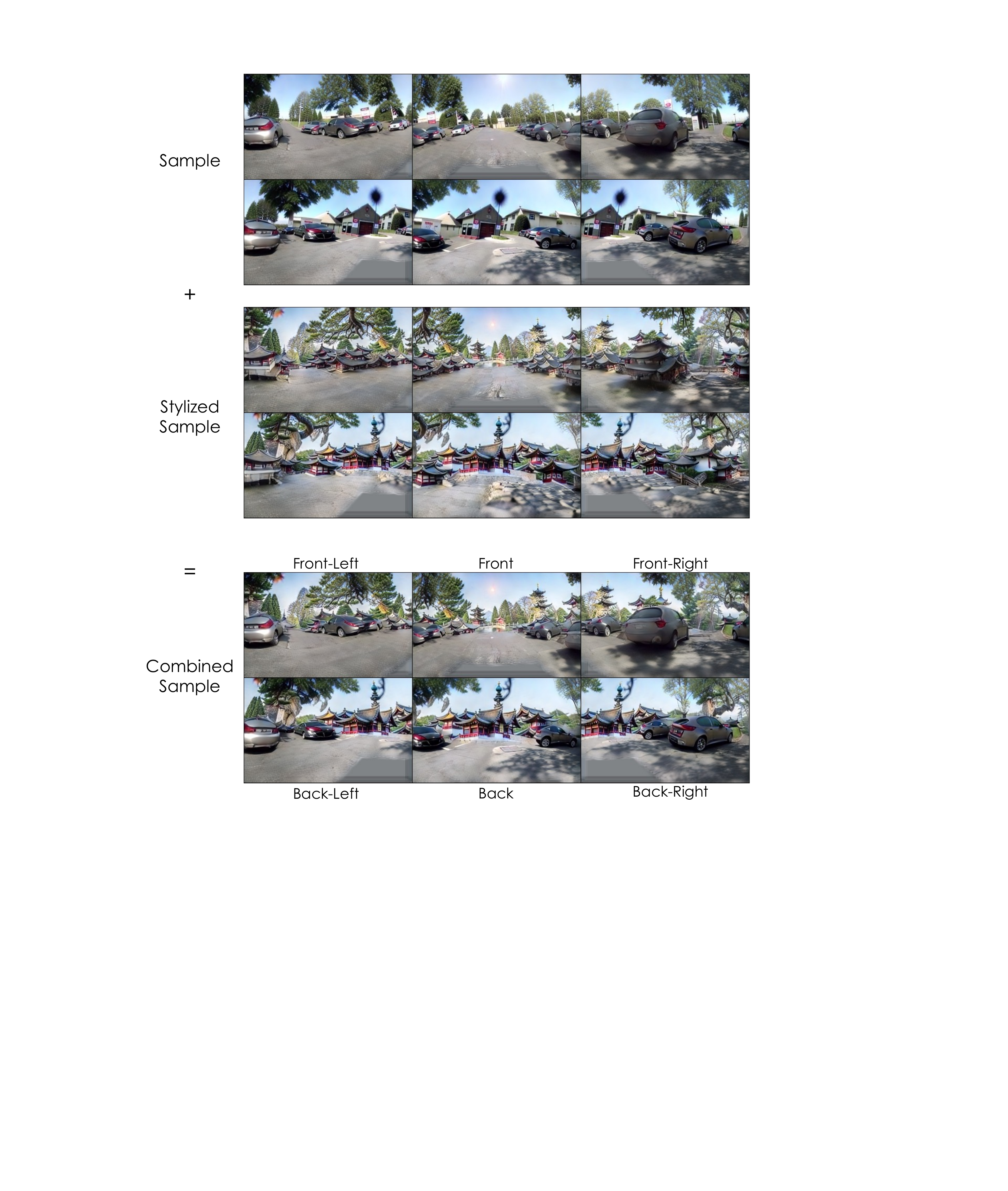}
   \caption{Combining voxels: we replace the center part of the stylized voxel with that of the sample at the top. 
    }
\label{fig:supp_style_combine}
\end{figure*}

\begin{figure*}[!thb]
  \centering
  \hspace{-5mm}
\includegraphics[width=0.83\textwidth]{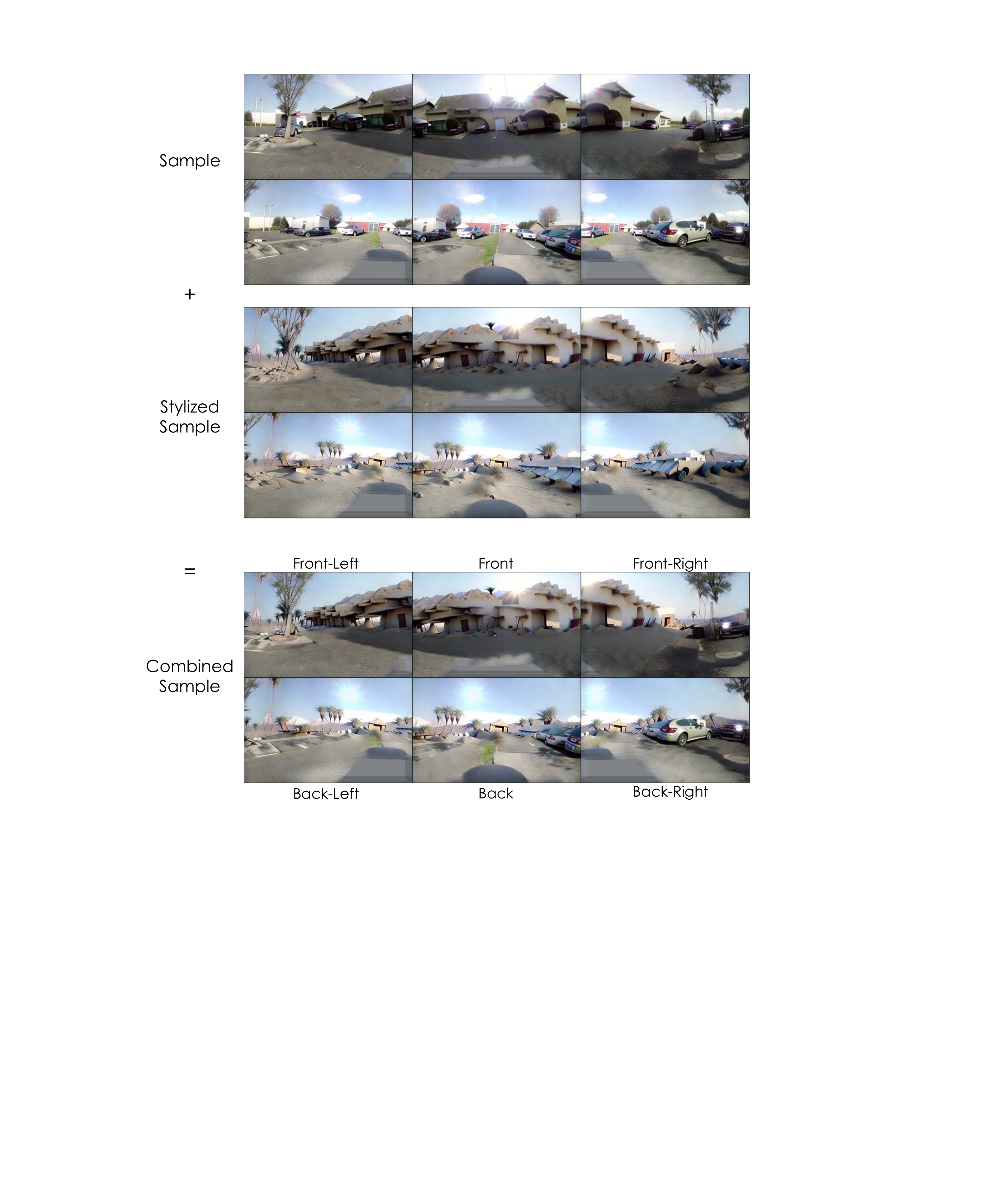}
   \caption{Combining voxels: we replace the center part of the stylized voxel with that of the sample at the top. 
    }
\label{fig:supp_style_combine2}
\end{figure*}

\begin{figure*}[!thb]
  \centering
  \hspace{-5mm}
\includegraphics[width=0.83\textwidth]{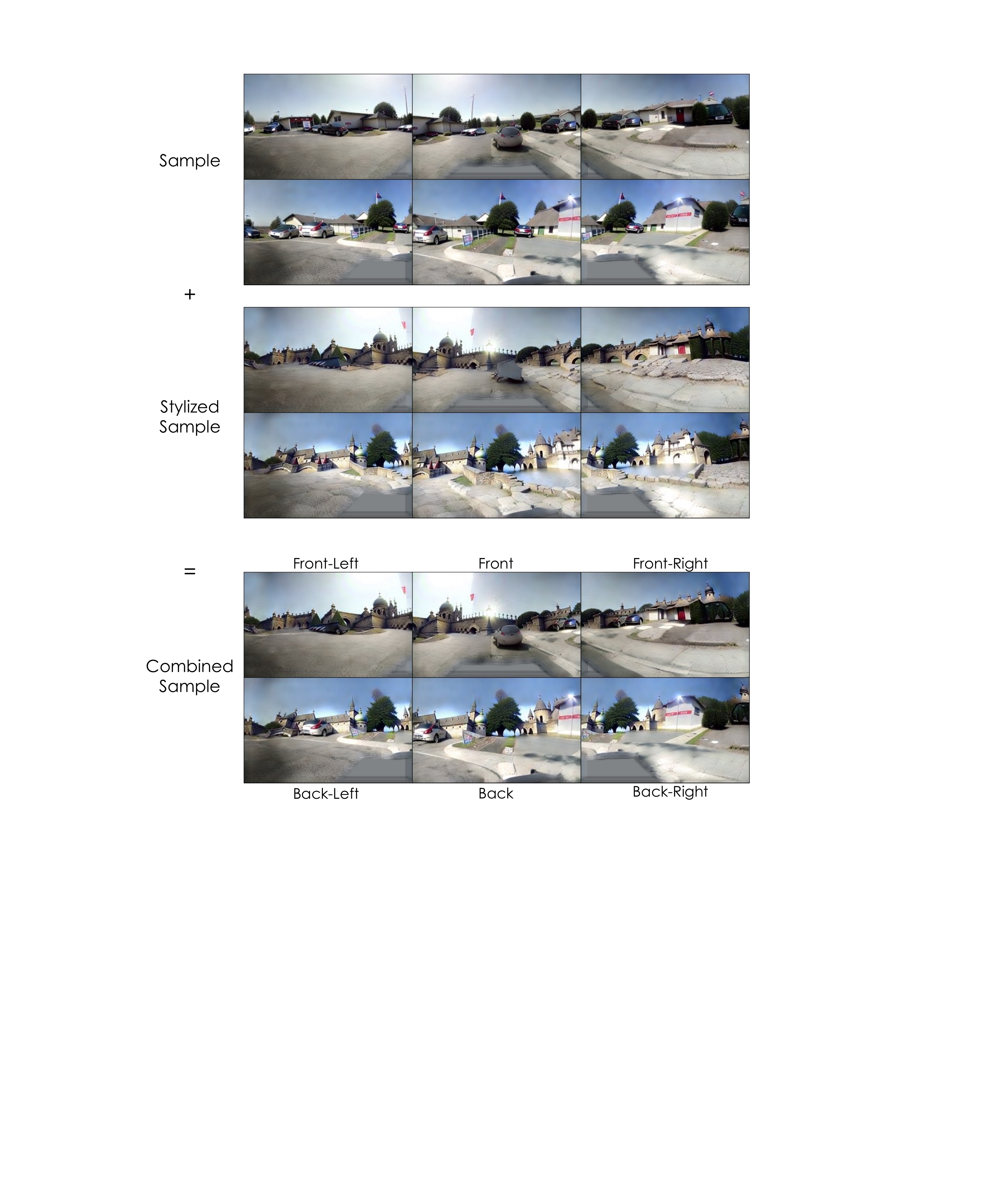}
   \caption{Combining voxels: we replace the center part of the stylized voxel with that of the sample at the top. 
    }
\label{fig:supp_style_combine3}
\end{figure*}

\begin{figure*}[!thb]
  \centering
\includegraphics[width=\textwidth]{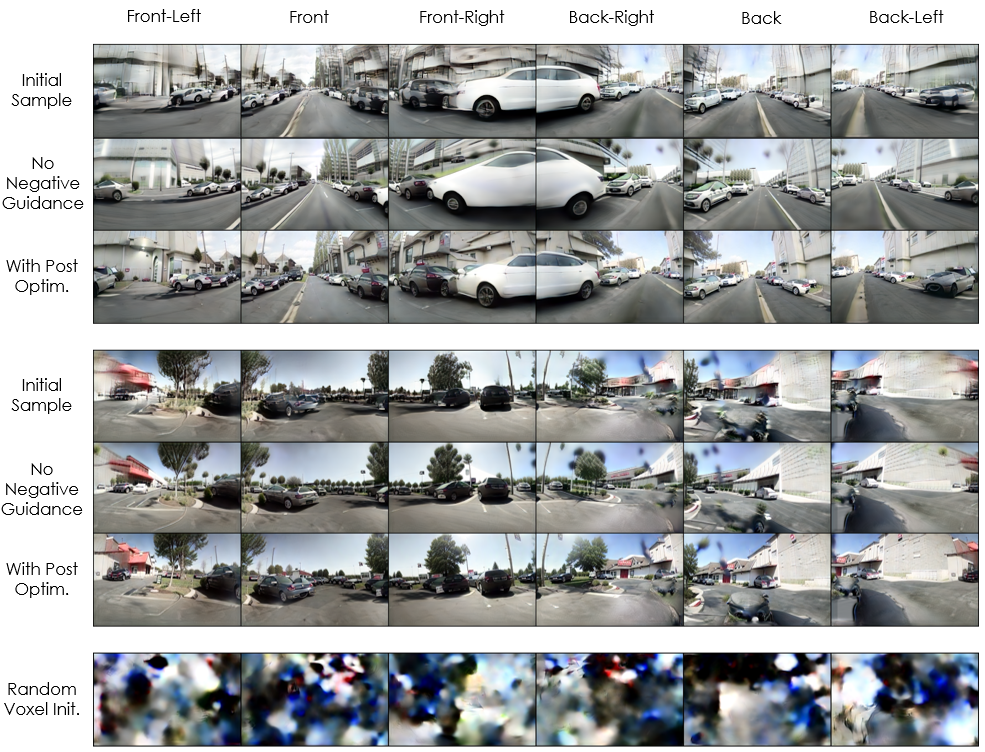}
   \caption{Ablating the post-optimization method. We show initial samples, samples optimized with classifier-free guidance and samples optimized with negative guidance for two scenes. Additionally, we show the result of post-optimization using a random gaussian intialization for the voxels.
    }
\label{fig:voxel_ablation}
\end{figure*}

We run SDS optimization with these modified conditioning vectors using the same procedure outlined in Section \ref{sec:post_opt}.
The stylization results can be seen in Figure \ref{fig:supp_style1}-\ref{fig:supp_style3}.
Moreover, as our neural fields are represented as voxel grids, we can easily combine different neural fields. 
In Figure~\ref{fig:supp_style_combine}-\ref{fig:supp_style_combine3}, we combine two sampled voxels by replacing the center region ($32\times 80\times 80$) of one voxel with the center region of the other one.
We qualitatively show the importance of having our initial voxel samples and the effect of negative guidance in Figure~\ref{fig:voxel_ablation}.

We note that the stylized scenes match the target style well, but do not perfectly preserve the content of the original scene (\eg the cars). 
For the scraping method, images for conditioning are randomly chosen and do not necessarily contain street scenes which could result in these semantic changes.
For the image translation method, we empirically found that parts of the translated scene with worse content preservation appeared differently when doing stylization with SDEdit multiple times on a single rendered image (\eg for lego stylization, cars contain different brick details and colors in each translation).
We hypothesize that doing SDS loss with this conditioning for thousands of iterations encourages the optimization to satisfy these multiple possible translations which results in blurring and a lack of content preservation in these regions.
Performing the post-optimization jointly with a reconstruction loss on images that preserves content and have the desired style (e.g. obtained through
the same img2img translation) could improve content preservation.


\clearpage

\subsection{Bird's Eye View Conditioned Synthesis}
\begin{figure*}[!thb]
  \centering
\includegraphics[width=0.7\textwidth]{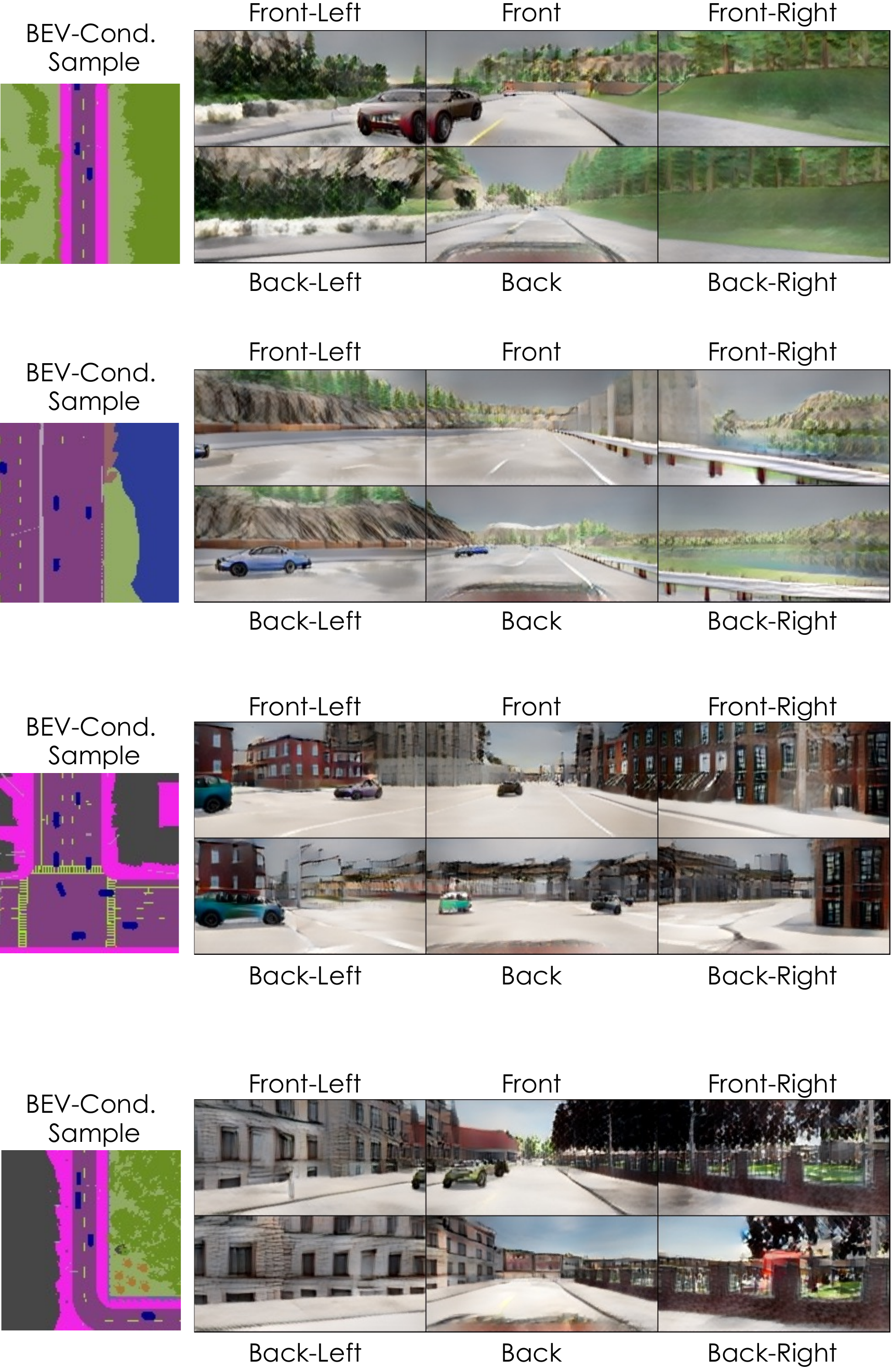}
   \caption{Additional results on Birds' Eye View Conditioned generation. In the BEV segmentation map, colors denote different region types: green - trees and vegetations, blue - water, grey - buildings, purple - road, pink - sidewalk,  dark blue - vehicles (note that the ego car is at the center and thus not visualized).
    }
\label{fig:bev}
\end{figure*}

We provide additional Bird's Eye View conditioned synthesis results in Figure~\ref{fig:bev}.

\clearpage

\subsection{Scene Editing}
\begin{figure*}[!thb]
  \centering
\includegraphics[width=0.8\textwidth]{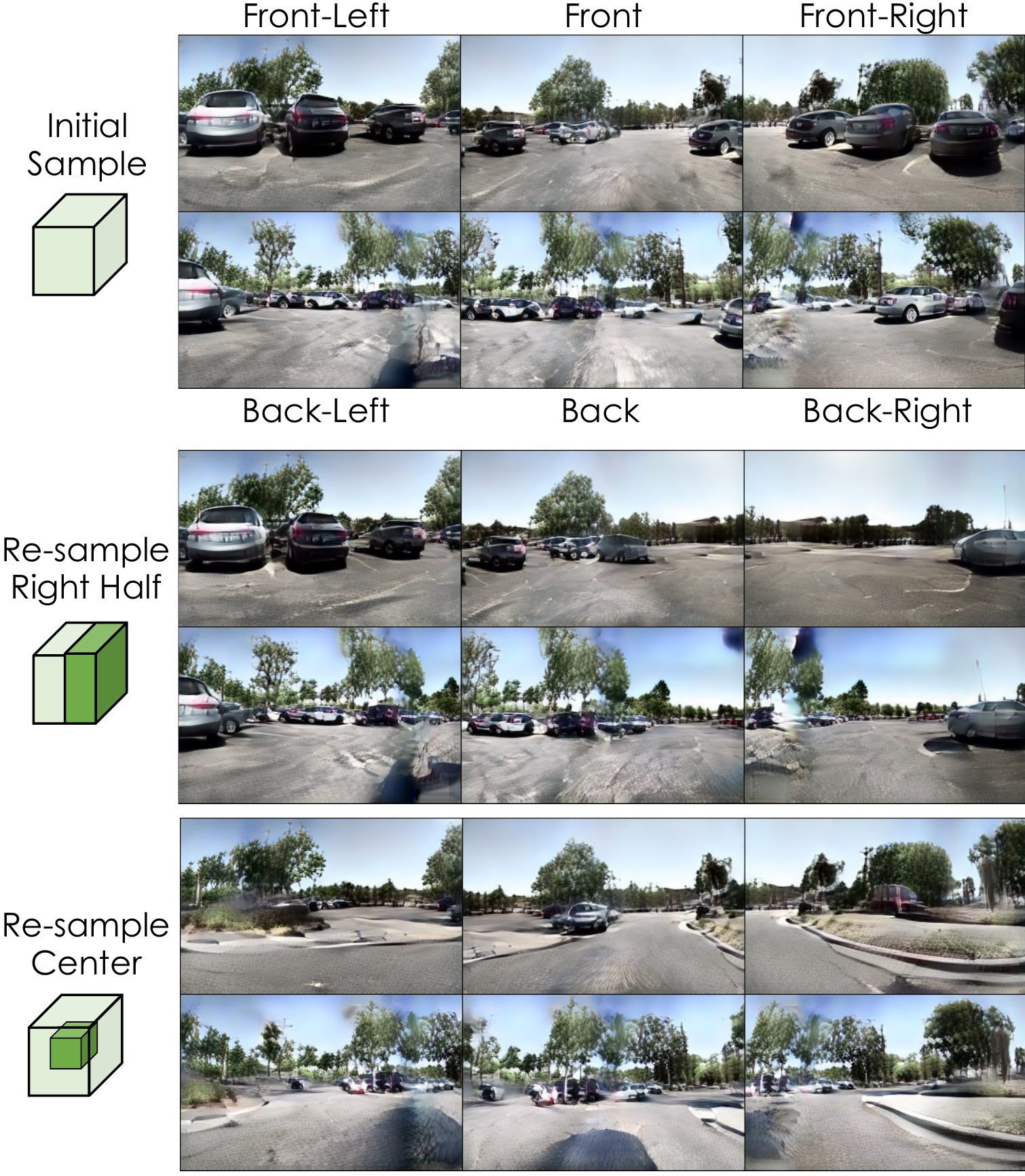}
   \caption{Additional results on scene editing by re-sampling. Given an initial sample, we edit the specified regions by re-sampling them with our model.
    }
\label{fig:editing1}
\end{figure*}

\begin{figure*}[!thb]
  \centering
\includegraphics[width=0.8\textwidth]{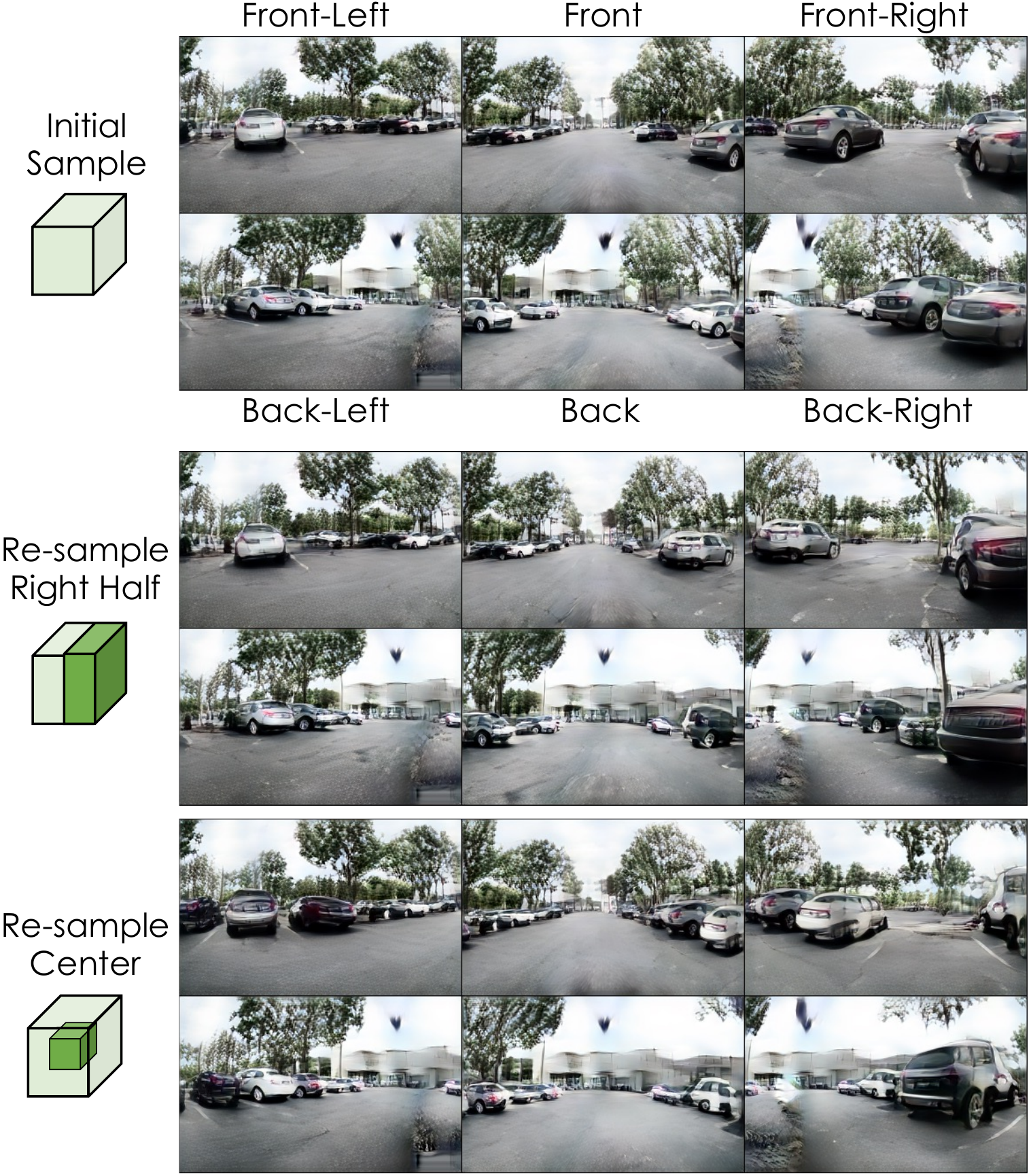}
   \caption{Additional results on scene editing by re-sampling. Given an initial sample, we edit the specified regions by re-sampling them with our model.
    }
\label{fig:editing2}
\end{figure*}

We provide additional scene editing results in Figure~\ref{fig:editing1} and \ref{fig:editing2}.

\clearpage

\subsection{Text-Guided Style Transfer}

\begin{figure*}[!thb]
  \centering
\includegraphics[width=0.8\textwidth]{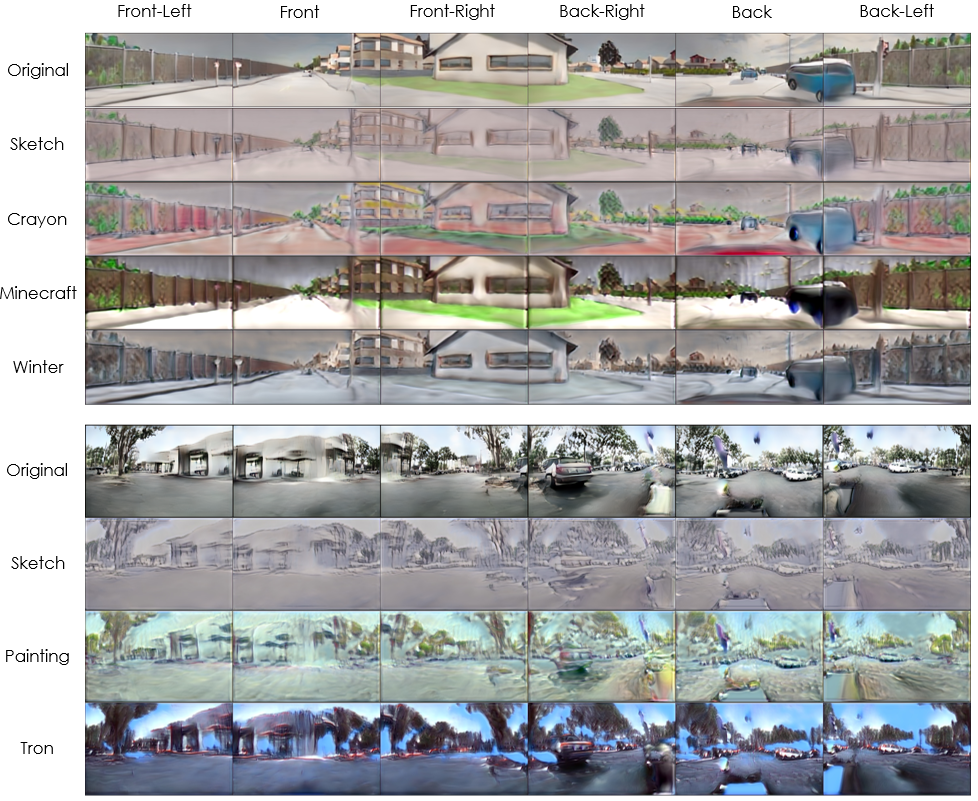}
   \caption{Text-guided style transfer results on an encoded Carla scene and a sampled AVD scene. Each result was obtained using a decoder finetuned by running CLIP directional loss with the specified style on a separate scene.
    }
\label{fig:stylegan_nada}
\end{figure*}

In addition to stylization using SDS loss, which is effective for high-quality large structural modifications, in Figure \ref{fig:stylegan_nada}, we show results of applying CLIP directional loss~\cite{gal2021stylegannada} to finetune our decoder for quick global style changes that generalize across scenes.
The target domain is expressed in natural language (\eg sketch of a city) and the source domain is either ``photo'' or ``photo of a city''.
We first obtain the update direction in CLIP space, $u_t = e_{target} - e_{source}$, where $e_{target}$ and $e_{source}$ are the CLIP text embeddings of the source and domain respectively. 
Then, we sample an encoded or sampled voxel, $V$, and initialize a frozen and trainable copy of our scene-autoencoder's decoder $D_f$ and $D_t$ respectively.
Additionally, we sample a set of camera paramaters $\{\kappa\}_{1 \dots N}$ from our dataset as our base poses.

At every iteration, we uniformly sample a translation offset in both the forwards and sideways directions between $-1$ and $1$ metres which we apply to the base poses to obtain jittered camera poses $\{\hat{\kappa}\}_{1 \dots N}$.
We render out images with the jittered poses using both the frozen and trainable decoders, obtaining $\hat{i}_f$ and $\hat{i}_t$ respectively.
We then obtain the current decoder's image update direction as $u_i = \hat{e}_{target} - \hat{e}_{source}$ where $e_{target}$ and $e_{source}$ are the CLIP image embeddings of $\hat{i}_f$ and $\hat{i}_t$ respectively.
The loss is then 1 minus the cosine similarity of $u_i$ and $u_t$, which is used to update only $D_t$.

We train using the Adam optimizer with learning rate set to $0.002$ and betas of $(0.9,0.999)$ between $20-100$  iterations, taking around a minute on a single V100 GPU.
We empirically found that while finetuning with CLIP directional loss is fast and training a domain-adapted model only requires optimizing on a single scene, SDS based stylization (Sections~\ref{sec:style_SDS}) produces much higher quality results.  

